\documentclass[10pt,twocolumn,letterpaper]{article}

\usepackage[pagenumbers]{wacv} % To force page numbers, e.g. for an arXiv version
\usepackage[accsupp]{axessibility}

\usepackage{graphicx}
\usepackage{amsmath}
\usepackage{amssymb}
\usepackage{booktabs}
\usepackage{graphicx}
\usepackage{subcaption}
\usepackage{etoolbox}
\usepackage{tikz}
\usepackage{comment}
\usepackage{amsmath,amssymb} % define this before the line numbering.
\usepackage{color}
\usepackage{threeparttable}
\usepackage{verbatim}
\usepackage{booktabs}
\usepackage{multirow}
\usepackage{xcolor}
\usepackage{pifont}
\usepackage{amsmath}
\usepackage{kotex}
\usepackage{times}
\usepackage{epsfig}
\usepackage{graphicx}
\usepackage{amsmath}
\usepackage{amssymb}

\newcommand{\fref}[1]{Fig.~\ref{#1}}
\newcommand{\sref}[1]{Sec.~\ref{#1}}
\newcommand{\tref}[1]{Table~\ref{#1}}

\newcommand{\eref}[1]{Eq.~\ref{#1}}

\usepackage[pagebackref,breaklinks,colorlinks]{hyperref}

\usepackage[capitalize]{cleveref}
\crefname{section}{Sec.}{Secs.}
\Crefname{section}{Section}{Sections}
\Crefname{table}{Table}{Tables}
\crefname{table}{Tab.}{Tabs.}

\def\wacvPaperID{352} % *** Enter the WACV Paper ID here
\def\confName{WACV}
\def\confYear{2024}

\begin{document}

%%%%%%%%% TITLE - PLEASE UPDATE
\title{Small Objects Matters in Weakly-supervised Semantic Segmentation}

\author{Cheolhyun Mun$^*$$^\dagger$\\
Samsung Research\\
Seoul, Korea\\
{\tt\small cheolhyunmun@yonsei.ac.kr}
% For a paper whose authors are all at the same institution,
% omit the following lines up until the closing ``}''.
% Additional authors and addresses can be added with ``\and'',
% just like the second author.
% To save space, use either the email address or home page, not both
\and
Sanghuk Lee$^*$$^\dagger$\\
SOCAR AI Research\\
Seoul, Korea\\
{\tt\small li-xh16@yonsei.ac.kr}
\and
Youngjung Uh \\
Yonsei University\\
Seoul, Korea\\
{\tt\small yj.uh@yonsei.ac.kr}
\and
Junsuk Choe\\
Sogang University\\
Seoul, Korea\\
{\tt\small jschoe@sogang.ac.kr}
\and
Hyeran Byun\\
Yonsei University\\
Seoul, Korea\\
{\tt\small hrbyun@yonsei.ac.kr}
}

\maketitle
\begin{abstract}
   Weakly-supervised semantic segmentation (WSSS) performs pixel-wise classification given only image-level labels for training. Despite the difficulty of this task, the research community has achieved promising results over the last five years. Still, current WSSS literature misses the detailed sense of how well the methods perform on different sizes of objects. Thus we propose a novel evaluation metric to provide a comprehensive assessment across different object sizes and collect a size-balanced evaluation set to complement PASCAL VOC. With these two gadgets, we reveal that the existing WSSS methods struggle in capturing small objects. Furthermore, we propose a size-balanced cross-entropy loss coupled with a proper training strategy. It generally improves existing WSSS methods as validated upon ten baselines on three different datasets.
\end{abstract}

%%%%%%%%% BODY TEXT
\section{Introduction} \label{sec:intro}

\begin{figure}[t!]
    \centering
    \begin{subfigure}[b]{\linewidth}
        \centering
        \includegraphics[width=\linewidth]{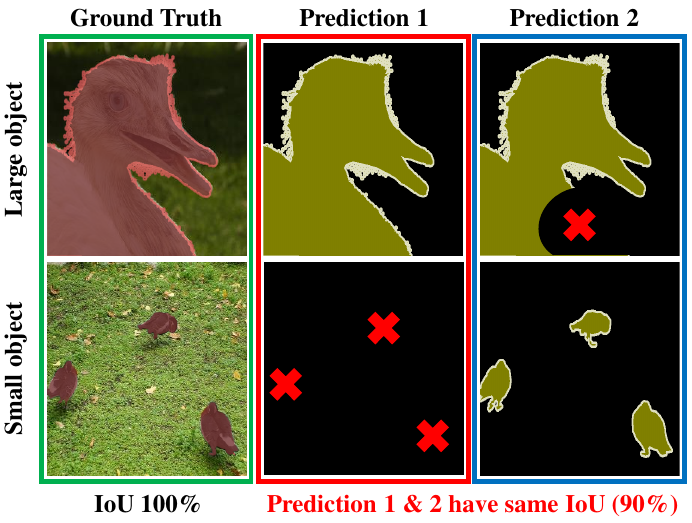}
        \caption{Large object domination problem in mIoU}
    \end{subfigure}
    \begin{subfigure}[b]{\linewidth}
        \centering
        \includegraphics[width=\linewidth]{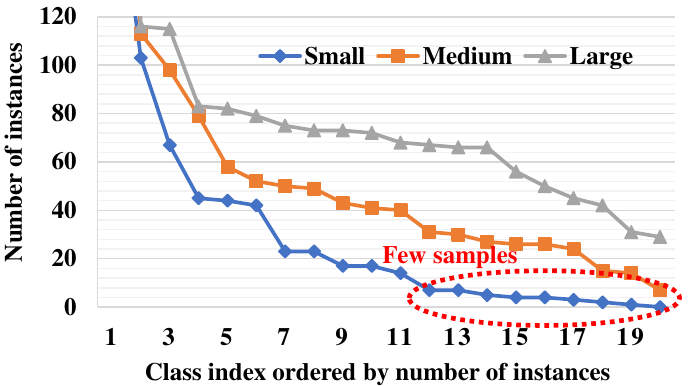}
        \caption{Imbalance problem in PASCAL VOC validation set}
    \end{subfigure}
    \caption{Problems of conventional metric and dataset. (a) Prediction 1 and 2 show the prediction for different cases which result in the same IoU scores. (b) Some classes of PASCAL VOC validation set suffer from a lack of small-sized objects. We sort the number of instances in descending order for each class per each size.}
    \label{fig:teaser}
\end{figure}

\def\thefootnote{*}\footnotetext{indicates equal contribution}
\def\thefootnote{$\dagger$}\footnotetext{The work was done while the author was at Yonsei University}

Recently, weakly-supervised learning (WSL) has been attracting attention because of its low-cost annotation. 
Among many tasks, weakly-supervised semantic segmentation (WSSS) methods learn to predict semantic segmentation masks given only weak labels such as image-level class labels for training.

To solve this problem, existing WSSS techniques generate pseudo segmentation masks from a classification network and then train a fully-supervised semantic segmentation model such as DeepLabV2~\cite{DeepLab}. To improve WSSS performances, most existing methods have focused on producing more accurate pseudo labels. With this strategy, WSSS performances have been greatly improved in the last five years~\cite{irn,rib,bbam,cda,bana,AMN,edam,Xie_2022_CVPR,nsrom,RCA}.

However, we lack a detailed sense of performance: do methods with high mIoU always better capture all the details? Interestingly, we observe that some methods with lower mIoU better capture small objects than others.
Although it is undoubtedly important that the segmentation model also correctly captures small objects, this limitation has not been well studied yet in WSSS literature.
How does each method behave in different types of environments? To answer this question, we address the limitations of the conventional metric, the dataset, and the training objective, and propose a complement thereby we anticipate WSSS techniques to become more complete and applicable to different needs.

\begin{figure}[t!]
    \centering
    \begin{subfigure}[b]{0.495\linewidth}
        \centering
        \includegraphics[width=\linewidth]{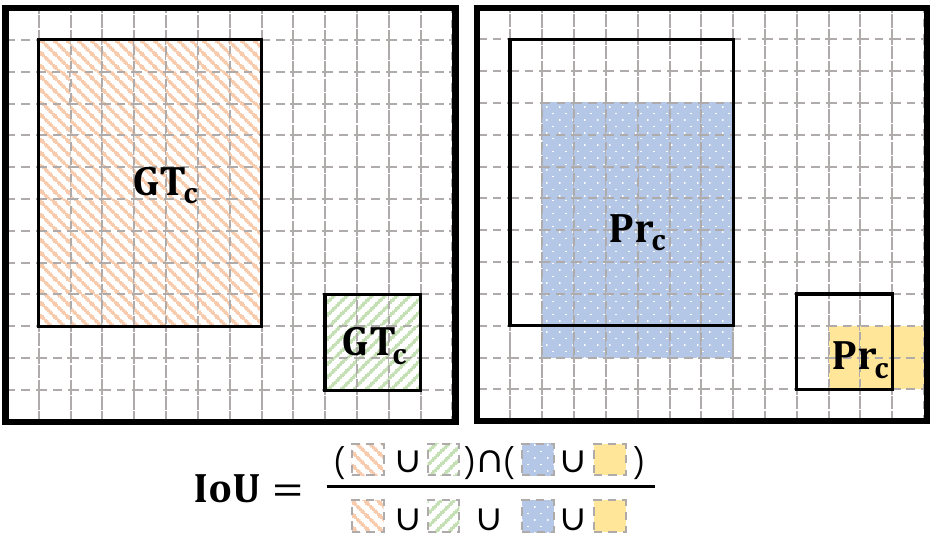}
        \caption{$\text{IoU}_{c}$ for \texttt{mIoU}}
    \label{fig:comparison_a}
    \end{subfigure}
    \begin{subfigure}[b]{0.495\linewidth}
        \centering
        \includegraphics[width=\linewidth]{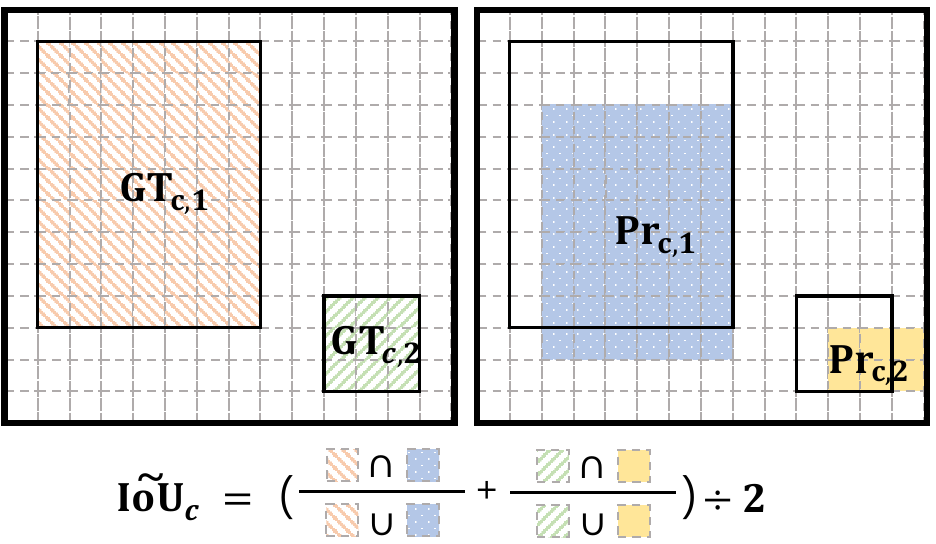}
        \caption{$\tilde{\text{IoU}}_{c}$ for \texttt{IA-mIoU}}
    \end{subfigure}
    \caption{Visual comparison of the computing process of $\text{IoU}_{c}$ for \texttt{mIoU} and $\tilde{\text{IoU}}_{c}$ for \texttt{IA-mIoU} regarding a class $c$}
    \label{fig:comparison}
\end{figure}

\noindent\textbf{Conventional metric (mIoU) and its pitfall.}
\texttt{mIoU} is mean of per-class IoUs where IoU is the intersection-over-union of the segmented objects. While an IoU is depicted with one predicted segment and one ground-truth segment, it pre-accumulates \textit{all} predicted pixels and \textit{all} ground-truth pixels in the entire dataset (\fref{fig:comparison} (a)).
\texttt{mIoU} has widely been used to measure the performance of different models in semantic segmentation. 

Despite of its usefulness in measuring the overall accuracy of segmentation predictions, \texttt{mIoU} does not account for the comprehensiveness of the predictions. As illustrated in~\fref{fig:teaser} (a), \textit{Prediction 1} and \textit{Prediction 2} have the same \texttt{IoU} score since they miss the same number of pixels. However, in \textit{Prediction 1}, the red cross marks indicate a complete failure in object segmentation, while \textit{Prediction 2} can be considered as minor errors.

\noindent\textbf{Conventional dataset.}
The PASCAL VOC 2012~\cite{Pascal} is the representative benchmark for WSSS. The problem is, however, the evaluation set of VOC has an imbalanced distribution in terms of object-size. 
\fref{fig:teaser} (b) shows the overall distribution for 20 classes of the VOC validation set per each size\footnote{Following MS COCO, we regard an instance as small if total number of pixels$<32\times32$, medium if the total number of pixels$<96\times96$, and large for the rest.}. Many classes fall short in the number of small objects.
Even with an ideal metric, we will never know how methods perform on small objects with few samples such as small birds.
Besides, we note that MS COCO~\cite{COCO}, another popular benchmark with 80 classes for WSSS, also suffers from imbalanced distribution. More information of dataset distribution is in the supplementary material.

\noindent\textbf{Training objective.} \label{motivation:loss}
Pixel-wise cross-entropy considers all individual pixels equally important by averaging. Thus the networks will consider small objects less important and lean toward large objects with many pixels. While the fully-supervised semantic segmentation methods have some remedy~\cite{FCN,focal}, WSSS literature has paid less attention to this problem. Existing works mostly focus on producing better pseudo masks to train the main segmentation network with the same pixel-wise cross-entropy.

\noindent\textbf{Our solutions.} In this paper, we suggest a way to address the above three limitations. First, we introduce a new evaluation metric for semantic segmentation, instance-aware mean intersection-over-union (\texttt{IA-mIoU}). It is important to accurately capture objects of all sizes to improve \texttt{IA-mIoU}. Next, we propose an evaluation dataset balanced in terms of object-size, PASCAL-B, which contains almost the same number of instances for each size, namely, large, medium, and small.
With our new benchmark and evaluation metric, we can correctly measure the performances of existing WSSS models in terms of object size. Specifically, we re-evaluate ten state-of-the-art methods~\cite{irn,rib,bbam,cda,bana,AMN,edam,Xie_2022_CVPR,nsrom,RCA} and observe interesting results; all evaluated methods struggle in capturing small objects.
Lastly, we propose a new loss function paired with a training strategy for segmentation models to balance the objective. Thorough experiments on three datasets demonstrate that our method achieves comprehensive performance boost on ten existing WSSS methods.
We believe that it will serve as a strong baseline to start with toward more comprehensive performance.
The code and the dataset will be publicly available for research community.

\section{Instance-aware mIoU} \label{sec:metric}
In this section, we explain how our metric addresses the limitations of \texttt{mIoU}. Then, we compare \texttt{mIoU} and our instance-aware mIoU (\texttt{IA-mIoU}) with the results of several corner cases.

\subsection{Definition of \texttt{IA-mIoU}.}
In \fref{fig:comparison} (a), we visualize the way of calculating $\text{IoU}_{c}$ of a class $c$, for \texttt{mIoU}. First, $\text{IoU}_{c}$ unions all pixels of ground-truth ($\text{GT}_{c}$) and prediction ($\text{Pr}_{c}$) respectively, and then calculates the intersection of them. During the process, it does not consider which instance each pixel belongs to. As a result, \texttt{mIoU} inherently does not provide a detailed sense of performance but provides coarse judgment.

To reflect the different importance of pixels, we suggest measuring the performance of each instance individually. We first split predictions and ground-truths of class $c$ into different instances \textit{i.e.,} $\text{Pr}_{c,1}$, $\text{Pr}_{c,2}$, $\text{GT}_{c,1}$ and $\text{GT}_{c,2}$ as shown in \fref{fig:comparison} (b). 
Then, we compute \texttt{IoU} scores $\text{IoU}_{c,i}$ for each instance $i$ and average them to obtain $\tilde{\text{IoU}}_{c}$ that is instance-aware IoU score of the class $c$:

\begin{gather} \label{eq3}
    \text{IoU}_{c,i}=\frac{{\text{Pr}}_{c,i}\cap {\text{GT}}_{c,i}}{{\text{Pr}}_{c,i} \cup  {\text{GT}}_{c,i}},~~~\tilde{\text{IoU}}_{c}= \frac{\sum_{i=0}^{T}\text{IoU}_{c,i}}{T},
\end{gather}
where $T$ is the total number of instances of the class $c$. Finally, we average the per-instance IoUs to compute instance-aware mIoU (\texttt{IA-mIoU}): 
 \begin{gather} \label{eq4}
    \texttt{IA-mIoU} = \frac{\sum_{c=0}^{N}\tilde{\text{IoU}}_{c}}{N}.
 \end{gather}
The following subsection describes how to split the predictions and ground-truths, and how to assign prediction instances to ground-truth instances.

\begin{figure}[t!]
    \centering
    \begin{subfigure}[b]{\linewidth}
        \centering
        \includegraphics[width=0.6\linewidth]{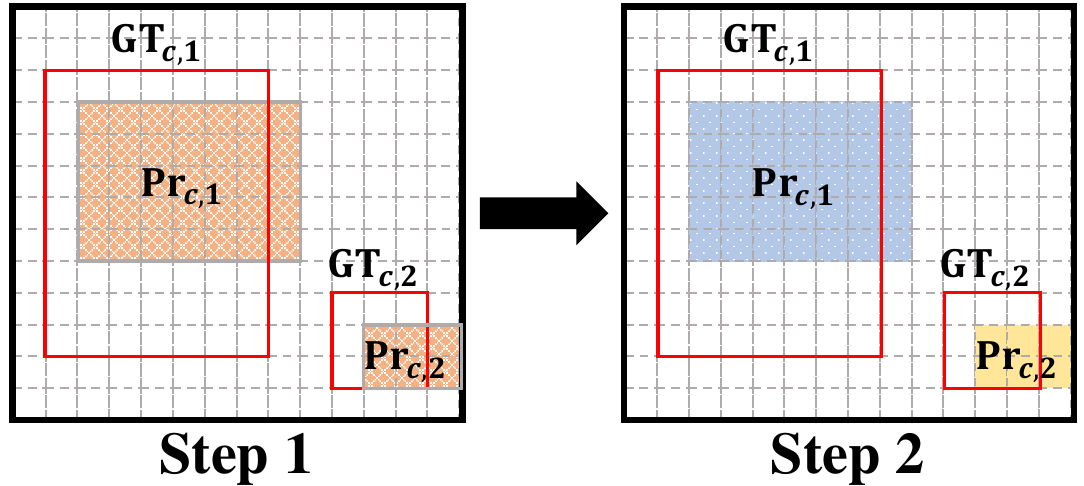}
        \caption{Case 1. One prediction segment covers one GT segment.}
    \end{subfigure}
    \begin{subfigure}[b]{\linewidth}
        \centering
        \includegraphics[width=\linewidth]{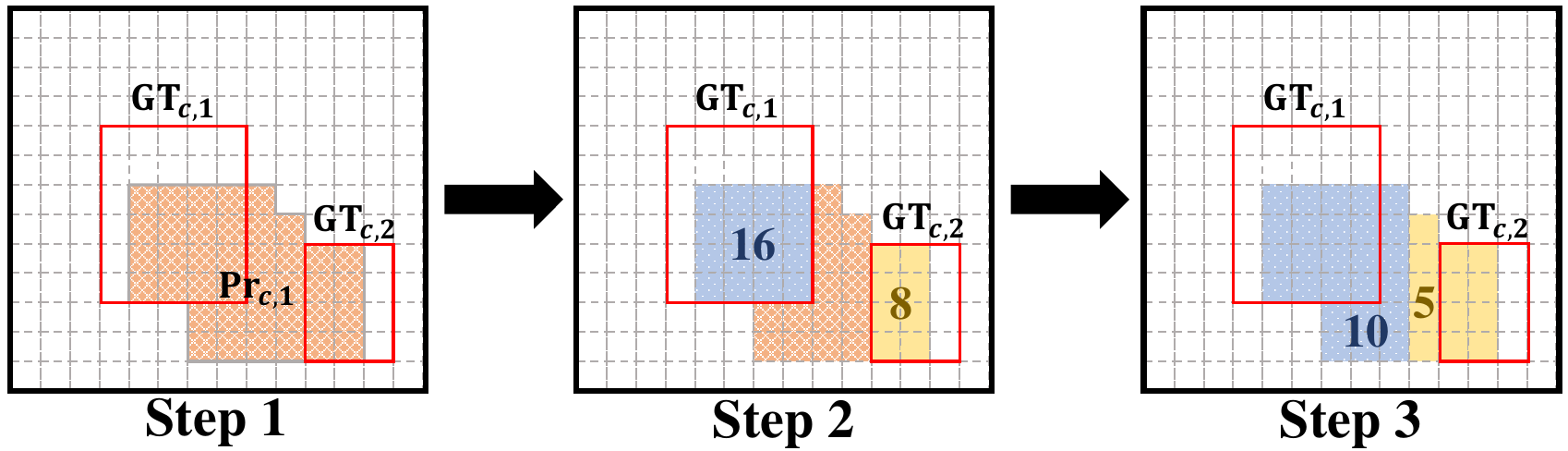}
        \caption{Case 2. One prediction segment touching multiple GT segments should be split. }
    \end{subfigure}
    \caption{Two cases for assigning predictions to the corresponding ground-truth instances. Pixels in color are the prediction and boxes with red lines are ground-truth instances. (a) When there is a one-to-one correspondence between prediction and ground-truth instance, each prediction is assigned to the corresponding ground-truth instance. (b) When there is a one-to-many correspondence between prediction and ground-truth instances, non-overlapping regions in step 2 (orange pixels with check pattern) distribute to each instance based on the ratio of blue and yellow pixels with dot pattern.}
    \label{fig:wnn}
\end{figure}

\subsection{Splitting and assigning instances}
Although we introduced the concept of instance, it does not exist in the segmentation task. Hence, we assume that the ground-truth segmentation masks can be either split into connected components (blobs) or split by additional instance annotation when available for evaluation. Please note that we introduce the instance labels only for more precise evaluation, not for training.

To fully utilize the instance masks for evaluation, we also have to split the predicted segments into blobs and assign them to overlapping ground-truth instances. There are three types of predictions for the model: 1) one prediction covers one object, 2) one prediction covers multiple objects simultaneously, and 3) prediction fails to cover any target instances. We consider only the first two cases because the last case has no overlapping region between prediction and ground-truth\footnote{False positives in a class $c$ do not contribute to $\tilde{\text{IoU}}_c$ but they decrease $\tilde{\text{IoU}}_\text{background}$.}

The procedure is illustrated in \fref{fig:wnn}. Both cases start from drawing contour lines from prediction for class $c$ ($\text{Pr}_{c}$) to get connected components ($\text{Pr}_{c,i}$).
The next step, however, is different for \textit{case 1} and \textit{case 2} since the former is a one-to-one correspondence relationship between $\text{Pr}_{c,i}$ and $\text{GT}_{c,i}$ and the latter is one-to-many.

For the \textit{case 1}, each connected component is assigned to overlapping target instance in the second step ($\text{Pr}_{c,1}\rightarrow\text{GT}_{c,1}$ and $\text{Pr}_{c,2}\rightarrow~\text{GT}_{c,2}$). Then, we can calculate the IoU per instance. On the other hand, for the \textit{case 2}, we have to split the connected component into multiple parts since it overlaps with multiple target instances. In other words, we have to distribute the non-overlapped area to each instances. To do this, we apply weighted clustering algorithm that if cluster ($i.e.,$ target instance) has more overlapped pixels than others, it takes larger unassigned regions. It has following advantages: 1) it does not favor or damage particular instances, 2) it is invariant to locations of the chosen pixels, and 3) it is less bias on the object size. 

This algorithm is implemented by adding two steps. We first assign the intersecting regions to the corresponding target instances and compute the ratio of the overlapping area ($i.e.,~\text{GT}_{c,1}\cap\text{Pr}_{c,1}:\text{GT}_{c,2}\cap\text{Pr}_{c,1}=16:8$) in the second step. In the final step, we distribute the remaining unassigned area to each target instance according to the ratio. The way of distribution of pixels can be not unique, but we focus on reasonable distributions of pixels based on instance size. For the multiple predictions and ground-truths, we would perform the same assignment process for each prediction and its corresponding ground-truth instances. This approach enables instance-aware metric in semantic segmentation tasks, even when the model does not provide instance-level predictions. In the next subsection, we design corner cases to compare the tendencies of \texttt{mIoU} and \texttt{IA-mIoU} clearly.

\begin{figure*}[t!]
    \centering
    \begin{subfigure}[b]{.48\linewidth}
        \centering
        \includegraphics[width=\linewidth]{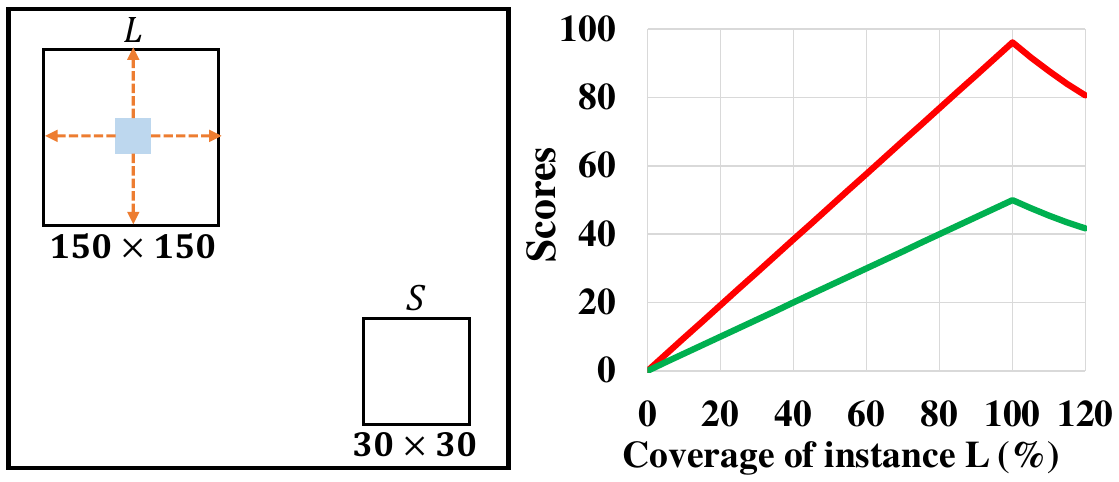}
        \caption{Case A}
    \end{subfigure}
    \begin{subfigure}[b]{.48\linewidth}
        \centering
        \includegraphics[width=\linewidth]{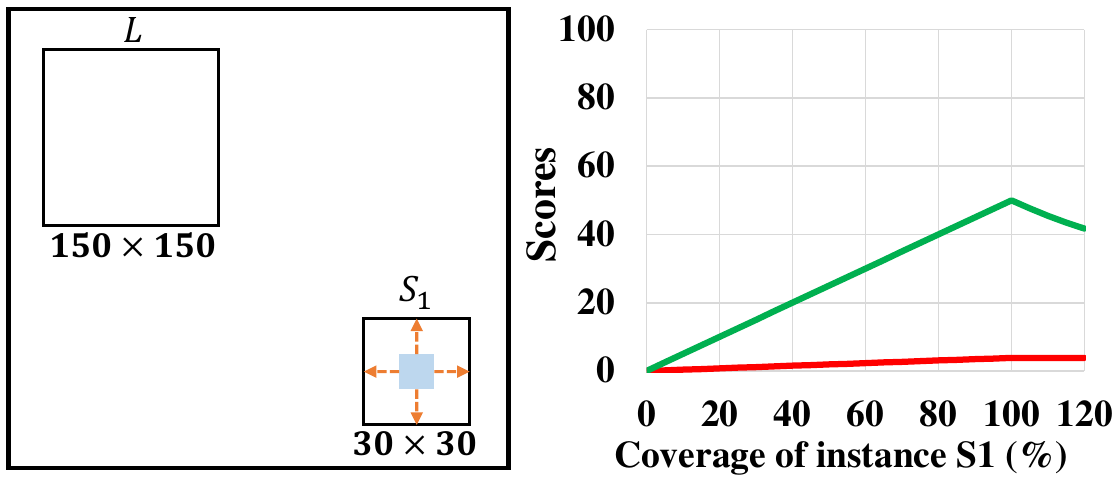}
        \caption{Case B}
    \end{subfigure}
    \\
        \begin{subfigure}[b]{.48\linewidth}
        \centering
        \includegraphics[width=\linewidth]{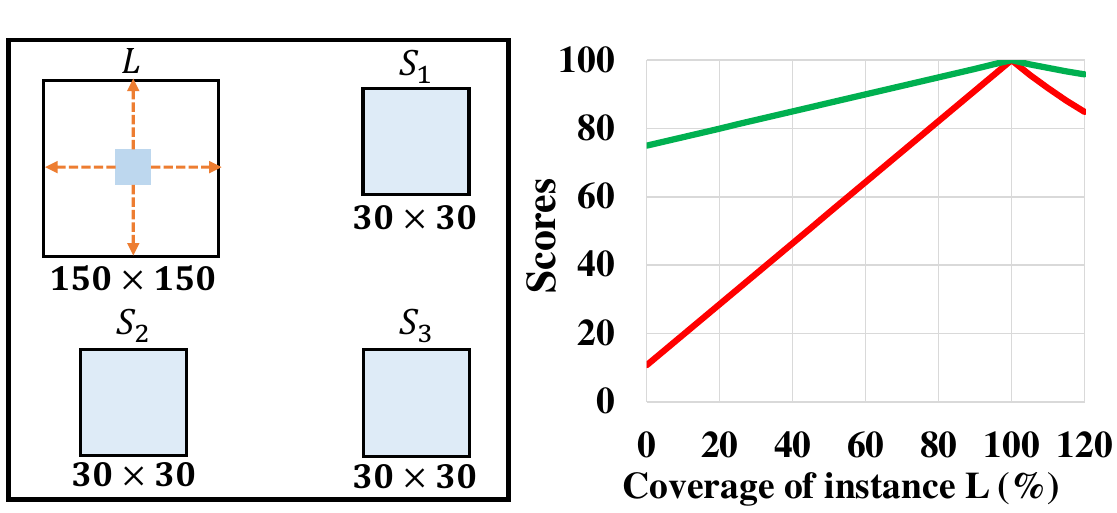}
        \caption{Case C}
    \end{subfigure}
    \begin{subfigure}[b]{.48\linewidth}
        \centering
        \includegraphics[width=\linewidth]{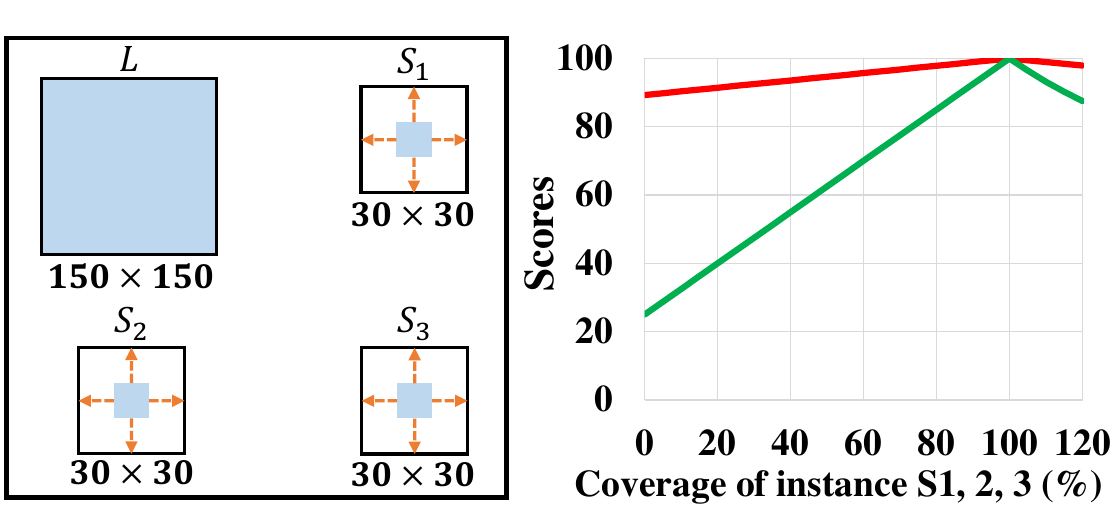}
        \caption{Case D}
    \end{subfigure}
    \includegraphics[width=0.35\linewidth]{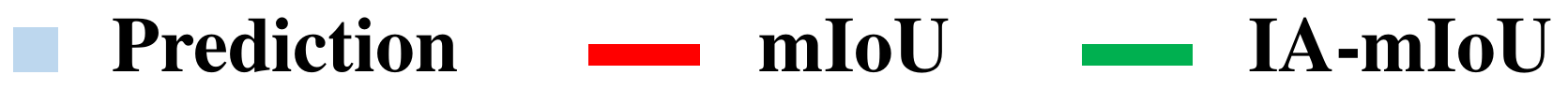}\vspace{-0.5em}
    \caption{Sensitivity to the size of instances on corner cases. We plot the behavior of \texttt{mIoU} and \texttt{IA-mIoU} as the prediction gradually grow to fill the ground-truth instance \textit{L} (or $\textit{S}_{1, 2, 3}$). Empty squares are uncovered ground-truth instances and sky blue squares are predictions. Gradual increase of the predictions is marked with orange dashed arrows.}
    \label{fig:corner_case}
\end{figure*}

\begin{figure}[t!]
  \centering
    \includegraphics[width=0.93\linewidth]{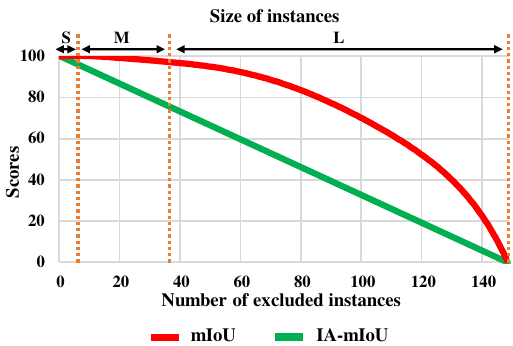}
  \caption{Corner case with real data. \texttt{mIoU} declines quickly as the size of instances gets larger while \texttt{IA-mIoU} drops consistently.} \label{fig:real_data}
\end{figure}

\subsection{Sensitivity analysis on corner cases.}
 We design four corner cases in~\fref{fig:corner_case}. We first set up small and large instances in an image, and then gradually expand the predictions to cover the assigned ground-truth instances. The outcomes show the limitation of \texttt{mIoU} more clearly: the prediction on a large object dominates the overall performance. The \texttt{mIoU} scores of \textit{case A} and \textit{C} increase exponentially with the improvement of prediction on a large object. On the contrary, the performances for \textit{case B} and \textit{D} barely change even though the predictions on small objects improve. Unlike the \texttt{mIoU}, our metric \texttt{IA-mIoU} steadily increases as the predictions fill the target instances regardless of the instance size. Furthermore, since we split the instances, we acquire more detailed sense of the performance according to their sizes ($i.e.,$ measuring only specific size of objects).
 
 In addition, \fref{fig:real_data} plots the behavior of \texttt{mIoU} and \texttt{IA-mIoU} in \textit{dog} class of the PASCAL VOC 2012 dataset. Starting from the perfect score, $i.e.,$ the prediction equals the ground-truth, we remove one instance at a time from the prediction starting with the smallest and progressing to the largest. \texttt{IA-mIoU} drops consistently, while \texttt{mIoU} barely decreases for small instances and rapidly decreases for large instances. We draw the red dashes in~\fref{fig:real_data} to distinguish the size of instances more clearly.

We hope that it would be beneficial for the community by providing a new comprehensive evaluation metric that can measure the semantic segmentation performance on small objects accurately.

\section{Dataset analysis and construction} \label{sec:dataset}

Imbalanced evaluation dataset may cripple the reliability of an evaluation protocol because the performance will vary due to the lack of samples. We believe that any objects with various sizes should not be undervalued because of their small number.

To tackle the imbalanced dataset issue, we suggest a new balanced benchmark dataset for evaluation. We construct PASCAL-B by collecting images and annotations from LVIS~\cite{lvis} and MS COCO~\cite{COCO} datasets which includes at least one of 20 categories\footnote{From MS COCO, we only collected images of ``potted-plant'' since LVIS does not have it.} of the PASCAL VOC classes. Then, we converted the annotations which do not belong to the 20 categories of the PASCAL VOC dataset into the background class. Among the remaining images, a few images have wrong annotations. Therefore, two computer vision experts (authors of this paper) manually filtered out such images for two weeks. Then, we randomly sampled images to ensure the balance over classes and object size distribution. In the end, PASCAL-B consists of 1,137 images with 20 classes. We give some representative images of the PASCAL-B dataset in the supplementary material.

As illustrated in~\fref{fig:average_dist} (b), our dataset is much more balanced in terms of classes and object-size distribution. Compared to PASCAL VOC, our PASCAL-B has fewer outliers, $i.e.,$ points in gray, and they do not have extremely large values. Also, PASCAL-B keeps a similar number of instances for each size while PSACAL VOC has more large or small instances.
In summary, a primary motivation for creating PASCAL-B was to address the issue of imbalanced evaluation datasets commonly encountered in semantic segmentation task. Existing benchmarks suffer from disparities in class or object size distributions, leading to skewed performance evaluations. PASCAL-B addresses this concern by meticulously constructing a dataset that features balanced classes and object sizes. Instead of replacing established benchmarks such as ADE20K~\cite{ADE20K}, COCO~\cite{COCO}, or Cityscapes~\cite{CityScapes}, PASCAL-B complements them by offering an alternative approach to assessment. For more details regarding the dataset, please refer to the supplementary material.
\begin{figure}[t!]
\centering
    \begin{subfigure}[b]{.45\linewidth}
        \centering
        \includegraphics[height=37mm,width=\linewidth]{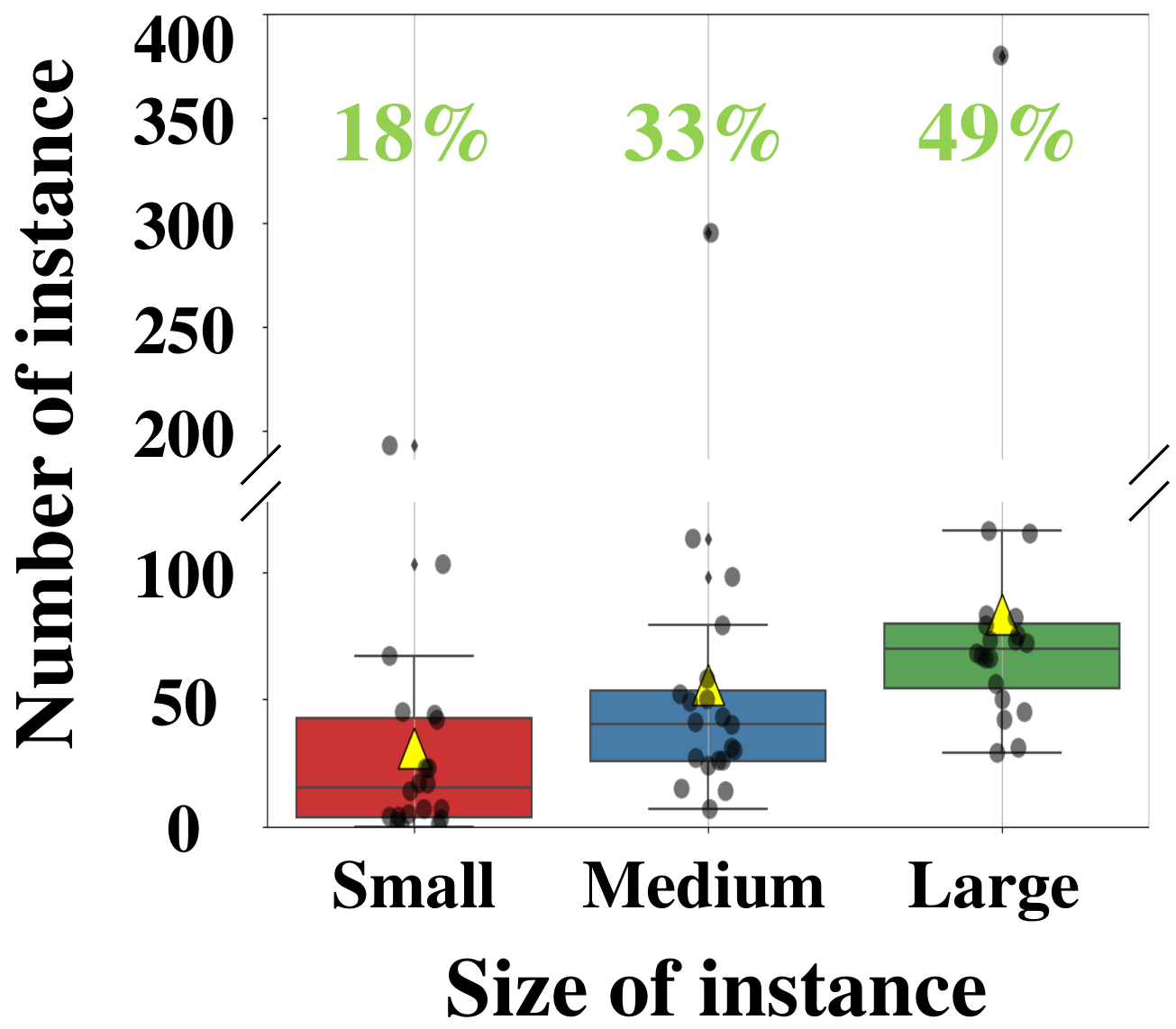}
        \caption{PASCAL VOC}
    \end{subfigure}
    \begin{subfigure}[b]{.45\linewidth}
        \centering
        \includegraphics[height=37mm,width=\linewidth]{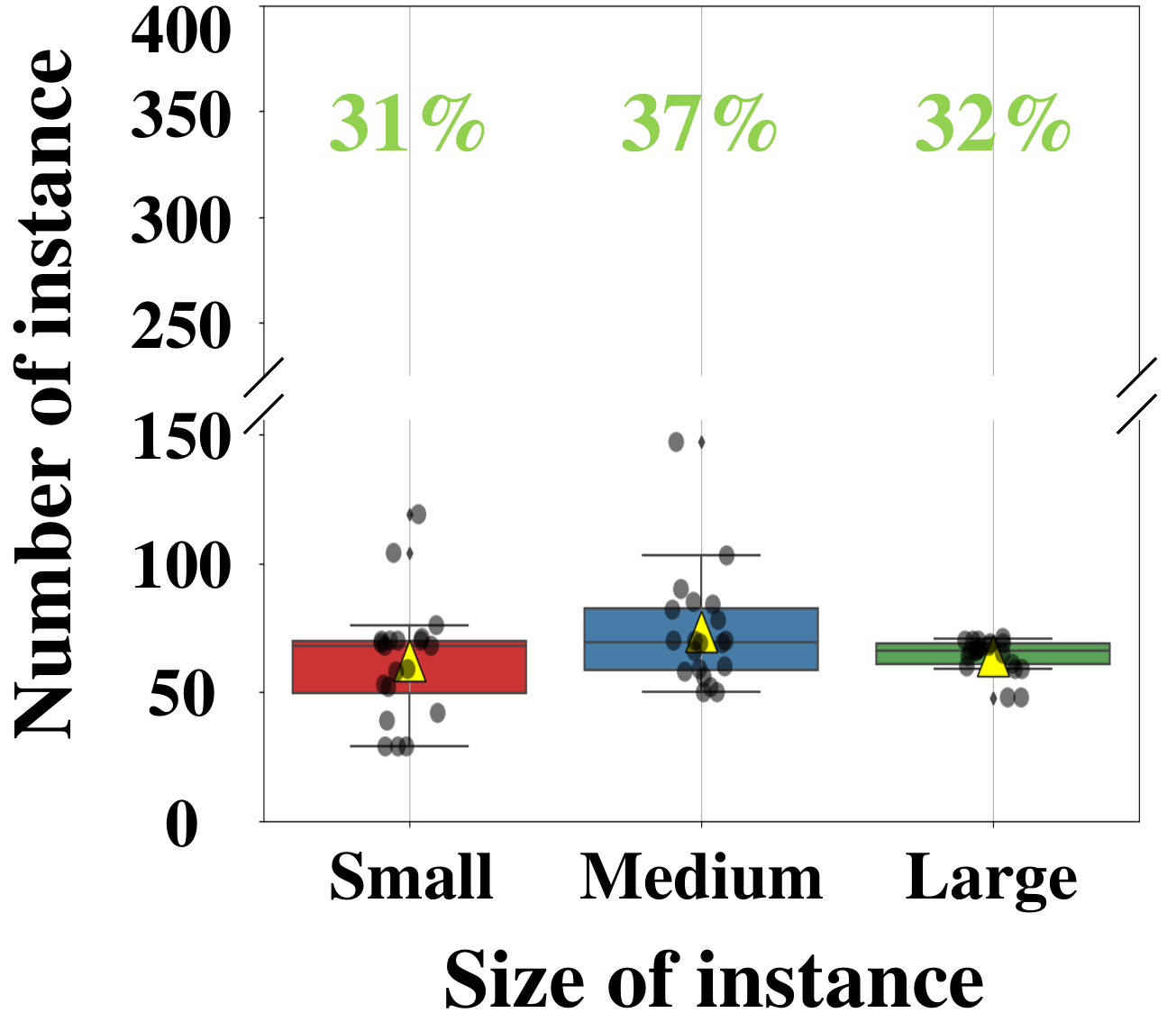}
        \caption{PASCAL-B}
    \end{subfigure}
  \caption{The distribution of validation set for each dataset: (a) PASCAL VOC and (b) PASCAL-B. We draw the mean ($i.e.,$ the triangle in yellow) and the variance over classes for each size of instances ($i.e.,$ \textit{small}, \textit{medium}, and \textit{large}). The point in gray indicates the number of instances for each class. On the top of each figure, we report the ratio of each size of instances to the total number of instances.}
  \label{fig:average_dist}
\end{figure}

\section{Methods} \label{sec:methods} 
\subsection{Evaluated WSSS methods} \label{sec:methods-sub1}
We evaluate ten existing methods under various weak-levels of supervision: bounding box supervision ($i.e.,$ BANA~\cite{bbam} and BBAM~\cite{bana}), saliency supervision ($i.e.,$ RCA~\cite{RCA}, EDAM~\cite{edam} and NS-ROM~\cite{nsrom}), natural language supervision ($i.e.,$ CLIM~\cite{Xie_2022_CVPR}), and image supervision ($i.e.,$ AMN~\cite{AMN}, RIB~\cite{rib}, CDA~\cite{cda}, and IRN~\cite{irn}). These methods follow the two-stage training pipeline of WSSS. In the first stage, they generate the pseudo masks by their methods. Then, they train a semantic segmentation network with the pseudo masks from the first stage. All the above methods except BANA~\cite{bana} only focus on stage 1 to produce the high-quality masks by refining the initial seed to improve the performance. A more detailed explanation for the above methods is in the supplementary material.

\subsection{Proposed loss function and training strategy} \label{sec:methods-sub2}
\begin{figure}
  \centering
    \includegraphics[width=\linewidth]{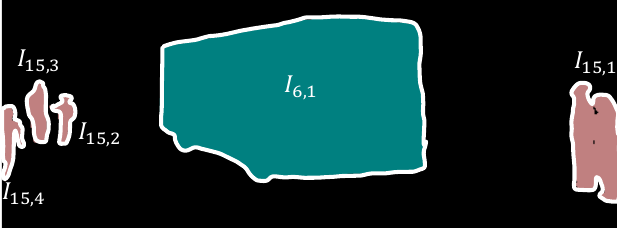}
  \caption{Example connected components for the loss function. $I_{c,k}$ is the \textit{k}-th connected components for \textit{c}-th class in an image.} \label{fig:loss_fig}
\end{figure}

To address the limitation of pixel-wise cross-entropy (CE) loss in~\sref{motivation:loss}, we propose a new loss function for a model to have the capacity of capturing small objects. We first give weights to each pixel according to the size of the object when computing the loss. Since the instance ground-truth masks are not available for training, we find all connected components for each class from pseudo ground-truth masks as in~\fref{fig:loss_fig}. Then, we get weight $w_{x,y}$ corresponding to a pixel $(x,y)$ as follows:
 \begin{gather} \label{eq6}
     w_{x,y}=
     \begin{cases}
        1, & \text{ if } {(x,y)} \in background, \\ 
        \min(\tau, \frac{\sum_{k=1}^{K}\text{S}_{c,k}}{\text{S}_{c,n}})
        , & \text{ otherwise } %(x,y) \in {\text{S}_{c,n}}
    \end{cases}
 \end{gather}
 \noindent where $\text{S}_{c,k}$ is the number of pixels in its connected component $I_{c,k}$, while $n$ is the index number of instance which pixel \text{$(x,y)$} is included. $K$ is the number of instances with $c$-th class in an image. Through~\eref{eq6}, we assign a larger weight to the pixels of the relatively small instance while preventing the value of weight from getting excessively large by setting up the upper limit $\tau$. Finally, we multiply weights to cross-entropy loss as in~\eref{eq7} and we call this loss function $L_{sw}$ as size-weighted cross-entropy loss.
  \begin{gather} \label{eq7}
     L_{sw} = -\frac{1}{H\times W}\sum_{c=1}^{C}\sum_{x=1}^{H}\sum_{y=1}^{W}Y_{c,x,y}w_{x,y}log(p_{c,x,y}),
 \end{gather}
 \noindent where $H$ and $W$ is the height and width of images, respectively, and $p_{c,x,y}$ is the probability to predict the class of the pixel $(x,y)$ as $c$.

Even though $L_{sw}$ can improve the ability of the model to catch small objects, there is a side effect that the model fails to learn extremely large instances with $L_{sw}$ during the whole training process. Therefore, we apply a new training strategy that adds a regularization term to~\eref{eq7} by introducing elastic weight consolidation (EWC) \cite{ewc}. EWC helps model to learn new tasks continually while preserving the information of previous tasks. Following the strategy of EWC, we also divide the training into two tasks. We first train a model using pixel-wise cross-entropy loss which is more beneficial to learn the large object as we analyze in \sref{motivation:loss}, and call this task as \textit{task A}. During the training for \textit{task A}, model updates the importance of parameters in Fisher information matrix. Then, for the new task, the model is fine-tuned by $L_{sw}$ and EWC helps to regularize the important parameter for the previous \textit{task A} based on the matrix. Thus, our final loss function $L_{sb}$, size-balanced cross-entropy loss, is defined as:

 \begin{gather} \label{eq8}
     L_{sb}=L_{sw}+\sum_{i}\frac{\lambda}{2}F_{i}(\theta_{i}-\theta_{A,i}^{*})^2,
 \end{gather}
 where $\theta_{i}$ and $\theta_{A,i}^{*}$ are $i$-th parameter for present task and \textit{task A}, respectively. $\lambda$ controls the importance of regularization and $F_{i}$ is the importance of parameter $i$ in the Fisher information matrix. With $L_{sb}$, a model can learn the new information for \textit{task B} (\textit{i.e.,} learning small objects) while maintaining the previous information from \textit{task A} (\textit{i.e.,} learning large objects).
 
\section{Experiments} \label{sec:experiment}
\subsection{Experimental setting} \label{sec:experiment-sub1}

\noindent\textbf{Dataset.} We evaluate each method on three datasets: PASCAL VOC~\cite{Pascal}, PASCAL-B, and MS COCO~\cite{COCO}. PASCAL VOC and PASCAL-B share the same training set though PASCAL-B is only designed for validation rather than training. PASCAL VOC and PASCAL-B consist of a similar number of images, 1,449 and 1,137, respectively.

\noindent \textbf{Evaluation metric.} We use \texttt{mIoU} and \texttt{IA-mIoU} to compare the performance of methods. Since our \texttt{IA-mIoU} can measure the small-sized instance only, we provide the $\texttt{IA}_{S}$ for the detailed performance of small objects.

\noindent \textbf{Implementation detail.} We generate pseudo masks for the segmentation networks using the official codes and strictly follow the setting provided in each paper~\cite{irn,rib,bbam,cda,bana,AMN,edam,Xie_2022_CVPR,nsrom,RCA}. Then, we use DeeplabV2 with ResNet-101~\cite{DeepLab} as segmentation networks. For more detail, please see the supplementary material. All the experiments were done by one GeForce RTX 3090 GPU for PASCAL VOC and two RTX 3090 GPUs for MS COCO.

\subsection{Quantitative results} \label{sec:experiment-sub2}
We evaluate nine baseline methods on PASCAL VOC and PASCAL-B, and three baseline methods on MS COCO. Furthermore, we demonstrate that our size-balanced cross-entropy loss function on the baseline methods results in better segmentation performance when compared to using the conventional cross-entropy (CE) loss.

\begin{table}[t!]
  \centering
  \fontsize{8}{10}\selectfont  
  \begin{threeparttable}  
    \begin{tabular}{c|c|c|cc}  
    \toprule
    \hline
    % \cmidrule(lr){2-5} \cmidrule(lr){6-9}\cmidrule(lr){10-13}
    Method&\texttt{Sup.}&\texttt{mIoU}&\texttt{IA-mIoU}&\texttt{IA$_{S}$}\cr  
    \hline
    IRN$^\ast$&$\mathcal{I}$&64.8&56.0&17.5\cr
    \hline
    CDA$^\ast$&$\mathcal{I}$&66.6&57.2&15.8\cr
    \hline
    AMN$^\ast$&$\mathcal{I}$&69.4&58.4&15.9\cr
    \hline
    CLIM$^\ast$&$\mathcal{I, L}$&68.9&57.4&14.0\cr

    \hline
    RCA$^\dagger$&$\mathcal{I, S}$&70.4&60.7&23.2\cr
    \hline
    EDAM$^\dagger$&$\mathcal{I, S}$&70.7&60.7&21.3\cr

    \hline
    NS-ROM$^\dagger$&$\mathcal{I, S}$&70.4&60.2&19.3\cr

    \hline
    BANA$^\dagger$&$\mathcal{I, B}$&72.6&59.6&14.7\cr
    \hline
    BBAM$^\ast$&$\mathcal{I, B}$&72.7&60.5&14.7\cr
    \hline
    \bottomrule  
    \end{tabular}  
    \end{threeparttable}
      \caption{Experimental results for PASCAL VOC. $\ast$ and $\dagger$ indicate that the segmentation model utilizes the ImageNet and COCO pretrained model respectively. $\mathcal{I}$, $\mathcal{S}$, $\mathcal{L}$ and $\mathcal{B}$ denotes the degree of supervision. $\mathcal{I}$: image-level supervision, $\mathcal{L}$: natural language supervision, $\mathcal{S}$: saliency supervision, and $\mathcal{B}$: bounding box supervision.} 
% \end{center}
    \label{tab:voc_table}  
\end{table}

\begin{table}[t!]
      \centering
      \fontsize{8}{10}\selectfont  
      \begin{threeparttable}  
        \begin{tabular}{c|c|c|cc}  
        \toprule
        \hline
        % \cmidrule(lr){2-5} \cmidrule(lr){6-9}\cmidrule(lr){10-13}
        Method&\texttt{Sup.}&\texttt{mIoU}&\texttt{IA-mIoU}&\texttt{IA$_{S}$}\cr  
        \hline
        DeepLabV2$^\ast$&$\mathcal{F}$&55.4&33.5&12.9\cr
        \hline
        RIB$^\ast$&$\mathcal{I}$&44.6&29.2&11.4\cr
        \hline
        IRN$^\ast$&$\mathcal{I}$&39.7&25.8&9.4\cr
        \hline
        \bottomrule  
        \end{tabular}  
        \end{threeparttable}
    % \end{center}
      \caption{Experimental results for MS COCO.} 
        \label{tab:coco_table}  
\end{table}

\begin{table}[b!]
  \centering
  \fontsize{8}{10}\selectfont  
  \begin{threeparttable}  
    \begin{tabular}{c|c|c|cc}  
    \toprule
    \hline
    % \cmidrule(lr){2-5} \cmidrule(lr){6-9}\cmidrule(lr){10-13}
    Method&\texttt{Sup.}&\texttt{mIoU}&\texttt{IA-mIoU}&\texttt{IA$_{S}$}\cr  
    \hline
    IRN$^\ast$&$\mathcal{I}$&56.1&41.0&15.8\cr
    \hline
    CDA$^\ast$&$\mathcal{I}$&57.5&41.4&13.4\cr
    \hline
    AMN$^\ast$&$\mathcal{I}$&58.5&41.1&13.9\cr
    \hline
    CLIM$^\ast$&$\mathcal{I, L}$&58.7&40.2&12.2\cr
    \hline
    RCA$^\ast$&$\mathcal{I, S}$&60.8&45.5&18.4\cr
    \hline
    EDAM$^\dagger$&$\mathcal{I, S}$&60.4&45.2&19.4\cr
    \hline
    NS-ROM$^\dagger$&$\mathcal{I, S}$&58.9&43.6&16.2\cr
    \hline
    BANA$^\dagger$&$\mathcal{I, B}$&61.9&41.1&14.0\cr
    \hline
    BBAM$^\ast$&$\mathcal{I, B}$&60.1&40.9&14.3\cr
    \hline
    \bottomrule  
    \end{tabular}
    \end{threeparttable}
      \caption{Experimental results for PASCAL-B.} 
    \label{tab:pascal-b_table}  
\end{table}

\paragraph{\texttt{mIoU} vs. \texttt{IA-mIoU}.}
\tref{tab:voc_table} compares the performances in \texttt{mIoU} and \texttt{IA-mIoU} on the PASCAL VOC dataset. Although the recent WSSS methods make impressive performance in the \texttt{mIoU} metric, we observe that the detailed scores measured by \texttt{IA-mIoU} are quite different. 
It is noteworthy that all WSSS methods get badly lower scores for small objects ($\texttt{IA}_{S}$) compared to overall scores. It indicates that WSSS methods struggle to capture the small instances accurately as we mentioned in~\sref{sec:intro}.

In particular, state-of-the-art techniques in terms of \texttt{mIoU} encounter more difficulty in capturing small objects compared to other methods. Consequently, they get lower \texttt{IA-mIoU} while getting the highest \texttt{mIoU}, since \texttt{IA-mIoU} reflects the scores of each instance equally but \texttt{mIoU} relatively neglects the small objects. This indicates \texttt{mIoU} fails to catch the detailed sense of performance on different sizes of objects.

We do the same experiments on the MS COCO dataset in~\tref{tab:coco_table}. According to the result of these experiments, we further demonstrate that existing WSSS methods struggle with small objects and it has been overlooked with \texttt{mIoU}.

\paragraph{PASCAL VOC vs. PASCAL-B.}
\tref{tab:pascal-b_table} compares the performances of models on our newly proposed benchmark, PASCAL-B. The models in \tref{tab:pascal-b_table} use the same checkpoint from~\tref{tab:voc_table} which are trained using the PASCAL VOC training set.

% \begin{table}[t!]
%   \centering
%   \fontsize{8}{10}\selectfont  
%   \begin{threeparttable}  
%     \begin{tabular}{c|c|c|cc}  
%     \toprule
%     \hline
%     % \cmidrule(lr){2-5} \cmidrule(lr){6-9}\cmidrule(lr){10-13}
%     Method&\texttt{Sup.}&\texttt{mIoU}&\texttt{IA-mIoU}&\texttt{IA$_{S}$}\cr  
%     \hline
%     IRN$^\ast$&$\mathcal{I}$&56.1&41.0&15.8\cr
%     \hline
%     CDA$^\ast$&$\mathcal{I}$&57.5&41.4&13.4\cr
%     \hline
%     CLIM$^\ast$&$\mathcal{I, L}$&58.7&40.2&12.2\cr
%     \hline
%     AMN$^\ast$&$\mathcal{I}$&58.5&41.1&13.9\cr
%     \hline
%     RCA$^\ast$&$\mathcal{I, S}$&60.8&45.5&18.4\cr
%     \hline
%     EDAM$^\dagger$&$\mathcal{I, S}$&60.4&45.2&19.4\cr
%     \hline
%     NS-ROM$^\dagger$&$\mathcal{I, S}$&58.9&43.6&16.2\cr
%     \hline
%     BANA$^\dagger$&$\mathcal{I, B}$&61.9&41.1&14.0\cr
%     \hline
%     BBAM$^\ast$&$\mathcal{I, B}$&60.1&40.9&14.3\cr
%     \hline
%     \bottomrule  
%     \end{tabular}
%     \end{threeparttable}
%       \caption{Experimental results for PASCAL-B.} 
%     \label{tab:pascal-b_table}  
% \end{table}

We argue that evaluating methods using imbalanced datasets can lead to biased scores, even with our proposed metric. To better evaluate the ability of models, it is essential to have a sufficient number of samples for evaluation per object-size and per class. However, the imbalance in the PASCAL VOC dataset makes it difficult to validate models since some classes have no small-sized objects, or there are only a few samples available. This lack of data for certain classes limits the opportunities for models to be evaluated on their performance, leading to potential biases in the evaluation process. On the other hand, we address this issue by constructing PASCAL-B which includes a sufficient number of samples for each object-size while keeping a balanced distribution across classes.

In this manner, the results in \tref{tab:pascal-b_table} with PASCAL-B provide better comprehensive assessment of WSSS methods compared to the scores in \tref{tab:voc_table}. When comparing the results of both tables, we observe that ranking order of WSSS methods is barely changed for \texttt{mIoU} and \texttt{IA-mIoU} in \tref{tab:voc_table} (Spearman’s rho: 0.79). On the other hand, it has totally changed in \tref{tab:pascal-b_table} with PASCAL-B dataset (Spearman’s rho: 0.38), which indicates that \texttt{IA-mIoU} scores with PASCAL-B evaluates the performance of models differently. We believe that the fundamental reason for this phenomenon lies in the discrepancy of distributions in terms of instance sizes between the two datasets. This suggests that \texttt{IA-mIoU} and PASCAL-B are both necessary to properly evaluate per-size performances.

\begin{figure}[t!]
\centering
    \begin{subfigure}[b]{.485\linewidth}
        \centering
        \includegraphics[width=\linewidth]{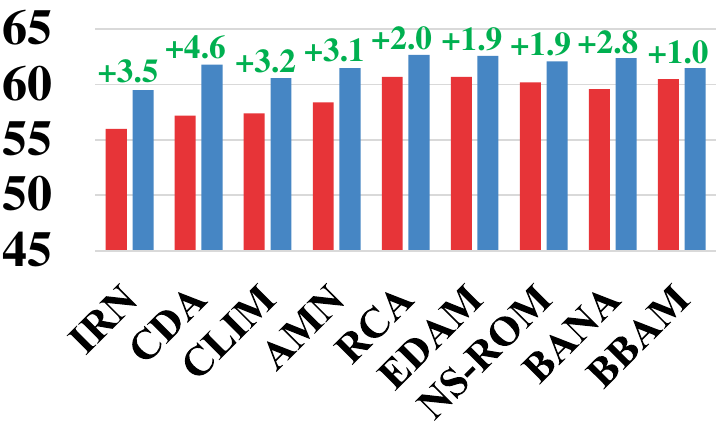}
        \caption{PASCAL VOC (\texttt{IA-mIoU})}
    \end{subfigure}
    \begin{subfigure}[b]{.485\linewidth}
        \centering
        \includegraphics[width=\linewidth]{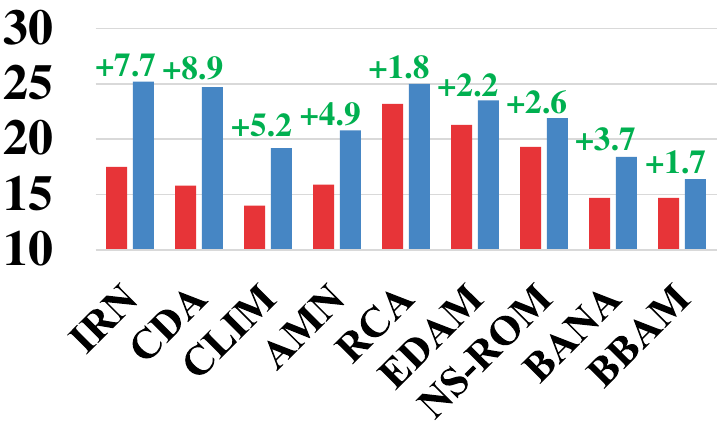}
        \caption{PASCAL VOC (\texttt{IA$_{S}$})}
    \end{subfigure}
    \\
    \begin{subfigure}[b]{.485\linewidth}
        \centering
        \includegraphics[width=\linewidth]{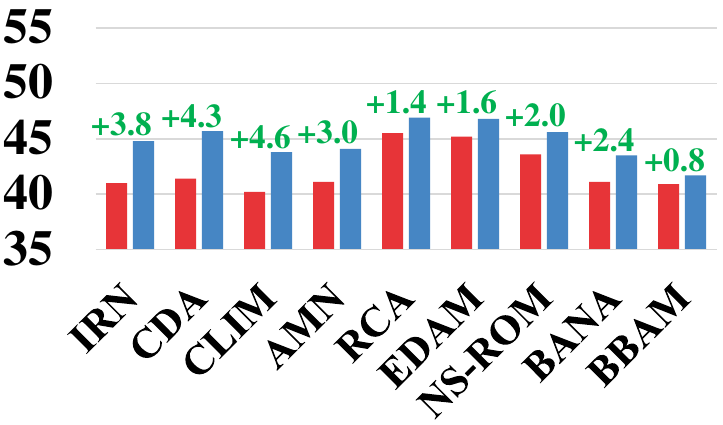}
        \caption{PASCAL B (\texttt{IA-mIoU})}
    \end{subfigure}
    \begin{subfigure}[b]{.485\linewidth}
        \centering
        \includegraphics[width=\linewidth]{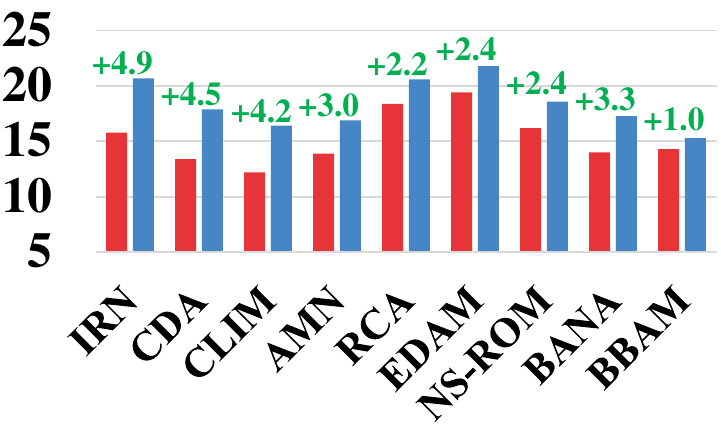}
        \caption{PASCAL B (\texttt{IA$_{S}$})}
    \end{subfigure}
    \\
    \begin{subfigure}[b]{.485\linewidth}
        \centering
        \includegraphics[width=\linewidth]{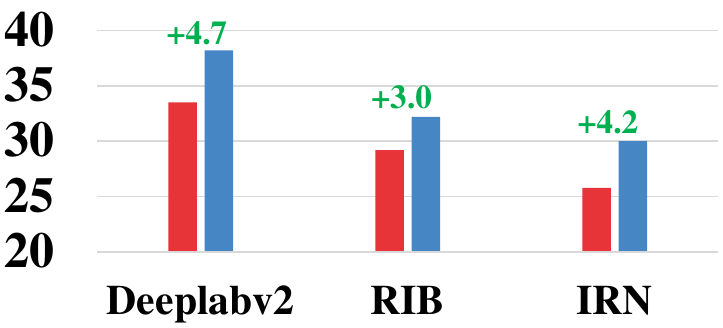}
        \caption{MS COCO (\texttt{IA-mIoU})}
    \end{subfigure}
    \begin{subfigure}[b]{.485\linewidth}
        \centering
        \includegraphics[width=\linewidth]{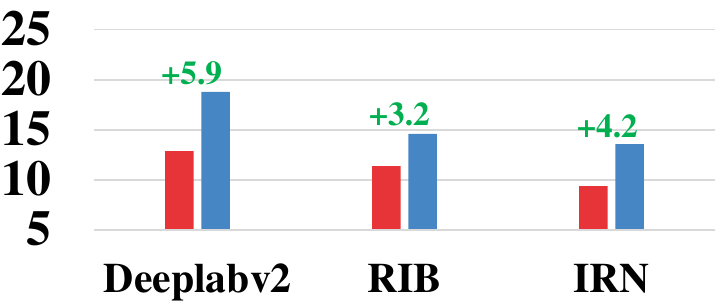}
        \caption{MS COCO (\texttt{IA$_{S}$})}
    \end{subfigure}
  \caption{Comparison of experimental results when applying CE loss (red bar) and Size-balanced CE loss (blue bar). We mark the increment above the bar (number with green color.)}
  \label{fig:compare_loss_func}
\end{figure}

\paragraph{CE loss vs. Size-balanced CE loss.}
Lastly, we verify the effectiveness of our proposed method, size-balanced cross-entropy loss function. As shown in~\fref{fig:compare_loss_func}, our method successfully boosts the performances for all models across three datasets. In particular, it enhances the ability of models to capture small instances. Across all datasets, we observe an increase of $\texttt{IA}_{S}$ scores ranging from 1.0 to 8.9. The changes of mIoU values, however, are relatively negligible, since the increase in performance of catching small instances has a little impact on mIoU as we explained in~\sref{sec:metric} (Out of 21 experiments, 18 have shown slight improvements in $\texttt{mIoU}$). In the supplementary material, we analyze qualitatively the experimental results according to the usage of our proposed loss function and provide more detailed values of performance gain. 

\subsection{Ablation study} \label{sec:experiment-sub3}
In this subsection, we demonstrate the effectiveness of each component of our loss function on the PASCAL VOC dataset with \texttt{mIoU} and \texttt{IA-mIoU}. In~\tref{tab:ablation_table}, we use a fully-supervised method, DeepLabV2~\cite{DeepLab} as our baseline model to observe performance gains by adding our components to the baseline.

\begin{table}[h!]
  \centering
  \fontsize{8}{10}\selectfont  
  \begin{threeparttable}  
    \begin{tabular}{c|c|cccc|c}  
    \toprule
    \hline
    % \cmidrule(lr){2-5} \cmidrule(lr){6-9}\cmidrule(lr){10-13}
    Method&\texttt{mIoU}&\texttt{IA-mIoU}&$\texttt{IA}_{S}$\cr 
    \hline
    DeepLabV2&77.8&65.8&18.8\cr
    with $L_{sw}$ &77.5&68.7&23.0\cr
    with $L_{sb}$&78.4&69.5&24.4\cr
    \hline
    \bottomrule  
    \end{tabular}  
    \end{threeparttable}
      \caption{Ablation study on each component of our loss function. $L_{sb}$: Add regularization to $L_{sw}$ using EWC.} 
       \label{tab:ablation_table}
\end{table} 

Applying only the size-weighted cross-entropy loss function $L_{sw}$ is powerful enough to gain notable improvements on small instances ($\texttt{IA}_{S}$) and \texttt{IA-mIoU} increases by 2.9 points. However, \texttt{mIoU} becomes slightly worse than the baseline. In other words, $L_{sw}$ alone does not ensure the same performance on the largest instances. 
$L_{sb}$ further boosts performance in all aspects by facilitating additional objective, covering small instances, while maintaining the previous objective, covering relatively large instances. Again, \texttt{IA-mIoU} enables detailed analyses by splitting the instances.
In short, introducing the size-balanced cross-entropy loss improves the performance on small instances and pairing EWC training strategy preserves the performance on large instances, resulting in overall improvement in both \texttt{mIoU} and \texttt{IA-mIoU}.

\section{Related Work} \label{sec:related_work}
\subsection{Weakly-supervised semantic segmentation}
Weakly-supervised semantic segmentation mainly adopts a two-stage pipeline: pseudo mask generation and training segmentation network. Most recent methods utilize Class Activation Maps (CAMs)~\cite{CAM} to generate a pseudo mask. However, CAMs have limitations in focusing on the most discriminative regions of the object or capturing frequently co-occurring background components. To solve this problem, lots of techniques have been proposed: adversarial erasing~\cite{ae,li2018tell,hou2018self,choe2020attention,cgnet,ecs-net}, seed growing~\cite{seed,huang2018weakly,rrm}, natural language supervision~\cite{Xie_2022_CVPR}, context decoupling~\cite{cda} and so on~\cite{bes,chang2020weakly,conta,advcam,rib}. Also, many methods~\cite{ficklenet,cian,oaa,eme,mcis,icd,auxsegnet,nsrom,edam} adopt a saliency supervision to refine the prediction map. It is usually utilized to enhance the result in a post-processing step by distinguishing the foreground and background of the object. Recently, Lee et al.~\cite{eps} try to make use of a saliency map during the training phase to maximize its potential. Besides, there are also some studies using a bounding box as a supervisory signal~\cite{BoxSup,BoxWSSS,box2seg,papandreou2015weakly,song2019box,bana,bbam} which is still cheaper than mask annotation. They achieve notable performance in WSSS since a bounding box label provides the exact location of all objects additionally. Our research, however, is interested in getting the better performance of models by improving the segmentation network in the second stage. Though few studies propose methods for segmentation networks, we suggest balanced training considering the size of instances in WSSS.

\label{sec:related_work1}
\subsection{Segmentation metrics} \label{sec:related_work2}
Here we briefly review the metrics for semantic segmentation. Pixel accuracy is the most basic metric for the task. It measures the accuracy for each class by computing the ratio of correctly predicted pixels of the class to all pixels of that class. The weakness of this metric is it does not consider false positives. Therefore, mean intersection-over-union (\texttt{mIoU}) replaces the pixel accuracy for semantic segmentation measures. It assesses the performance of models by calculating prediction masks intersection ground-truth masks over prediction masks union ground-truth masks. The \texttt{mIoU} compensates for the shortcoming of pixel accuracy by taking account of false positive. Nonetheless, as we analyze it in the next section, it still suffers from a size imbalance problem. Besides, various metrics~\cite{csurka2004good,kohli2009robust,Cordts2016Cityscapes,cheng2021boundary} are also investigated. Cordts et al.~\cite{Cordts2016Cityscapes} point out the inherent bias of the traditional IoU measure towards larger instances. They proposed instance-level IoU which focuses on adjusting pixel contributions based on instance sizes and class-averaged instance sizes, aiming to refine mIoU. However, our metric IA-mIoU evaluates each instance individually by segmenting predictions into instances, providing a comprehensive assessment that is not influenced by instance size.

\section{Conclusion} \label{sec:conclusion}
\subsection{Contributions}
In this paper, we focus on the comprehensive assessment and improvement of weakly-supervised semantic segmentation (WSSS) by proposing a novel metric, dataset, and loss function with an appropriate training strategy. First, we uncover the overlooked issue related to small-sized instances due to the conventional metric (\texttt{mIoU}). To address this, we design the instance-aware mIoU (\texttt{IA-mIoU}) to measure the performance of models more precisely regardless of object-size. Moreover, we point out the imbalance problem in benchmarks of WSSS and introduce a well-balanced dataset for evaluation, PASCAL-B. Lastly, we propose the size-balanced cross-entropy loss to compensate for the imbalance problem of pixel-wise cross-entropy loss. We show the effectiveness of our loss function on ten WSSS methods over three datasets measured by \texttt{mIoU} and \texttt{IA-mIoU}.

\subsection{Limitations}
Our findings can be applied to fully-supervised semantic segmentation methods. However, due to limited computing power, we were unable to utilize more recent FSSS models and evaluate them with datasets such as ADE20K~\cite{ADE20K} and Cityscapes~\cite{CityScapes}. Nevertheless, we hope that our study can serve as inspiration for other researchers who have the necessary resources to explore these avenues further.

\section{Acknowledgment}
This research was supported by the National
Research Foundation of Korea grant funded by the Korean government (MSIT) (No. 2022R1A2B5B02001467)

%%%%%%%%% REFERENCES
{\small
\bibliographystyle{ieee_fullname}
\bibliography{ms}
}

\end{document}

%%%%%%%%% TITLE - PLEASE UPDATE
\title{(Supplementary materials) Small Objects Matters in Weakly-supervised Semantic Segmentation}

\maketitle

We provide the following supplementary materials in this appendix:
\begin{itemize}
    \item In~\sref{sup_sec:dataset detail}, we illustrate the distribution of each dataset (i.e., PASCAL VOC, MS COCO, and PASCAL-B) and the procedure of building PASCAL-B dataset thoroughly.
    \item In~\sref{sup_sec:evaluated method}, we briefly explain each models which used for evaluation.
    % \item In~\sref{sup_sec:fully}, we demonstrate the effectiveness of our proposed metric, dataset, and loss function with fully-supervised methods.
    \item In~\sref{sup_sec:implementation detail}, we describe the implementation detail of each method we use.
    \item In~\sref{sup_sec:ewc}, we give a concise explanation of elastic weight consolidation~\cite{ewc}.
    \item In~\sref{sup_sec:fully}, we demonstrate the effectiveness of our proposed metric, dataset, and loss function with fully-supervised methods.
    \item \sloppy In~\sref{sup_sec:qualitative result}, we provide the qualitative results of each method on three datasets: PASCAL VOC, MS COCO, and PASCAL-B.
\end{itemize}
\section{Dataset details} \label{sup_sec:dataset detail}
\subsection{Number of instances per class per size}
\begin{figure}[t!]
    \centering
    \begin{subfigure}[b]{\linewidth}
        \centering
        \includegraphics[height=4mm]{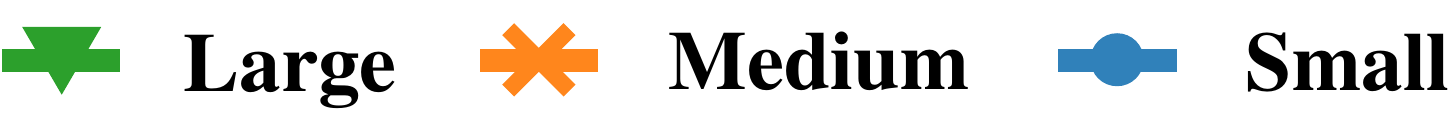}
    \end{subfigure}
    \begin{subfigure}[b]{\linewidth}
        \centering
        \includegraphics[width=\linewidth]{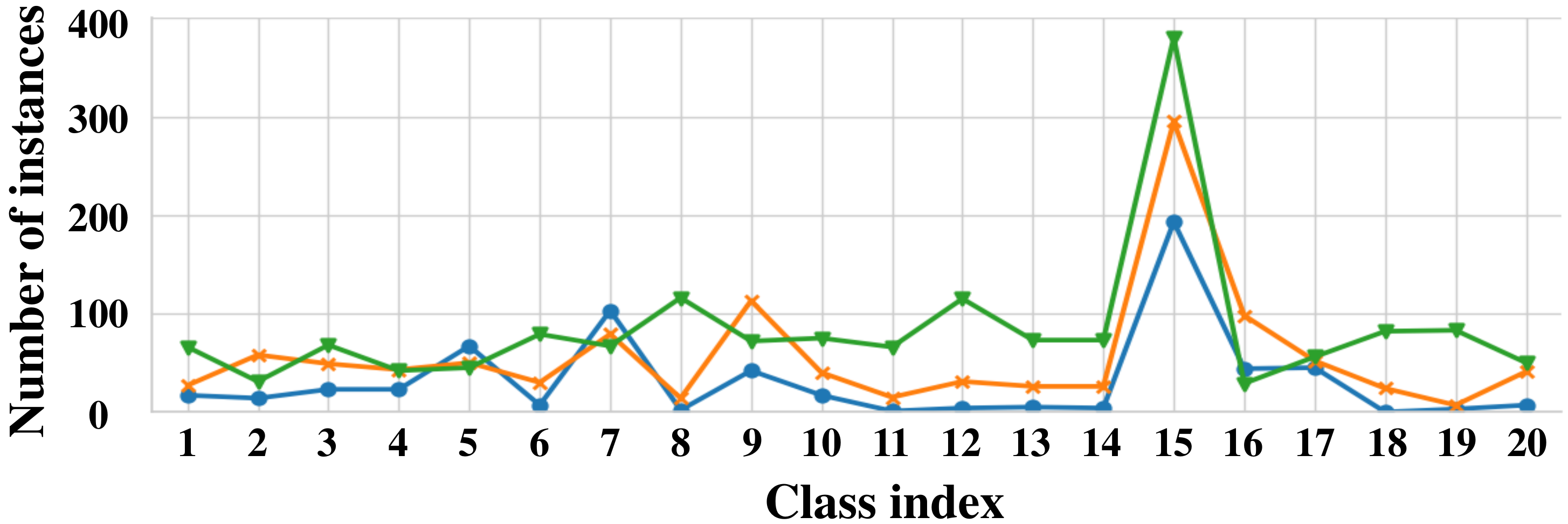}
        \caption{PASCAL VOC}
    \end{subfigure}
    \begin{subfigure}[b]{\linewidth}
        \centering
        \includegraphics[width=\linewidth]{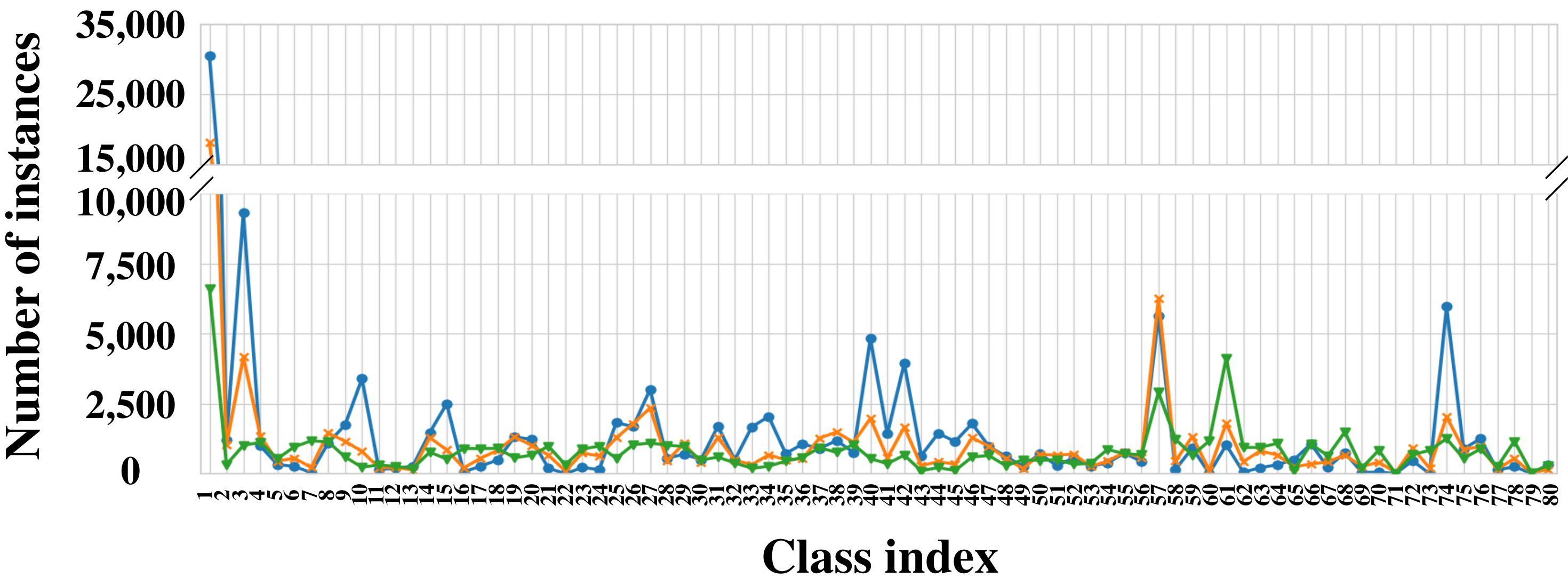}
        \caption{MS COCO}
    \end{subfigure}
    \begin{subfigure}[b]{\linewidth}
        \centering
        \includegraphics[width=\linewidth]{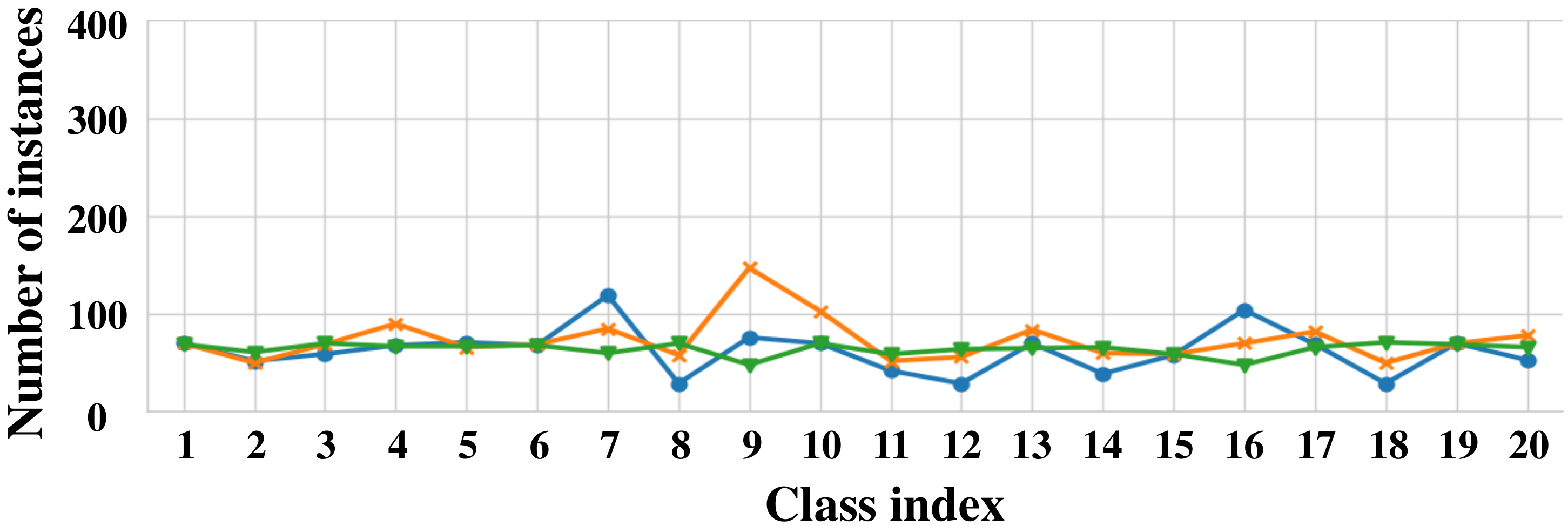}
        \caption{PASCAL-B}
    \end{subfigure}
    \caption{Dataset distribution. We plot the number of instances of each class by size.}
    \label{fig:sup_dataset}
\end{figure}

\fref{fig:sup_dataset} shows the per-class per-size distribution of validation set for each dataset in detail. 
As shown in \fref{fig:sup_dataset}(a), PASCAL VOC 2012~\cite{Pascal} suffers from an imbalance problem in terms of class and size of instances.
In particular, it has too many instances for the person class (i.e., 15th class) compared to the other classes. Some classes even do not have small instances.
For PASCAL VOC, large instances account for 50\% of the total number of instances while small instances only take 18.2\%.

Secondly, MS COCO~\cite{COCO} also has a serious class imbalance problem with some categories (\fref{fig:sup_dataset}(b)).
Additionally, it has imbalanced distribution in terms of instance size though the amount is less than PASCAL VOC. As in~\tref{tab:sup_dataset}, the number of small instances makes up about 43.7\% of the total instances while that of large instances is only 24.3\%. 

Different from these two datasets, PASCAL-B is the more balanced dataset.
~\fref{fig:sup_dataset} (c) illustrates that our dataset alleviates the problems of class and size imbalance. In other words, PASCAL-B does not have the case that a specific class has too many instances and it has similar number of instances for all sizes as shown in~\tref{tab:sup_dataset}.

\begin{table}[h!]
  \centering
  \fontsize{7}{10}\selectfont  
  \begin{threeparttable}  
    \begin{tabular}{c|c|c|c}  
    \toprule
    \hline
    % \cmidrule(lr){2-5} \cmidrule(lr){6-9}\cmidrule(lr){10-13}
    Instance size&PASCAL VOC&MS COCO&PASCAL-B\cr 
    \hline
    Large&1,668 (\textbf{\textcolor[rgb]{1,0,0}{49.0\%}})&65,407 (24.3\%)&1,283 (32.1\%)\cr
    \hline
    Medium&1,118 (32.8\%)&86,469 (32.1\%)&1,468 (\textbf{\textcolor[rgb]{1,0,0}{36.7\%}})\cr
    \hline
    Small&621 (18.2\%)&117,789 (\textbf{\textcolor[rgb]{1,0,0}{43.7\%}})&1,245 (31.2\%)\cr
    \hline
    Total&3,407&269,665&3,996\cr
    \hline
    \bottomrule  
    \end{tabular}  
    \end{threeparttable}
    \caption{The number of instances by size for each dataset.} 
    \label{tab:sup_dataset}
% \end{center}
\end{table}

\subsection{Process of constructing new dataset} 
\begin{figure}[t!]
    \centering
        \includegraphics[width=\linewidth]{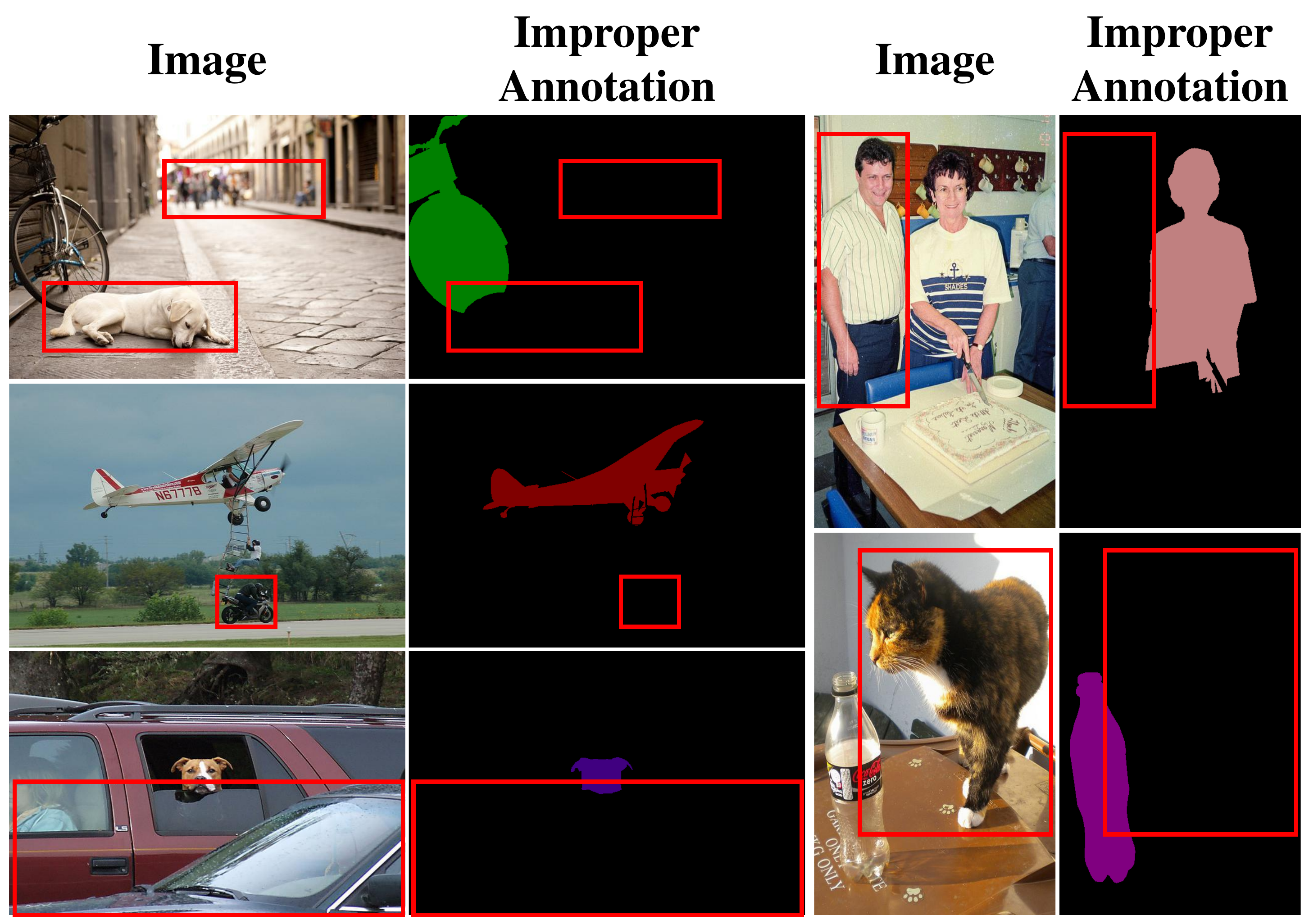}
        \caption{Example images with improper annotations. Red bounding boxes indicate missing annotations.}
    \label{fig:our_wrong}
\end{figure}

\begin{figure*}[h!]
    \centering
        \includegraphics[width=0.95\linewidth]{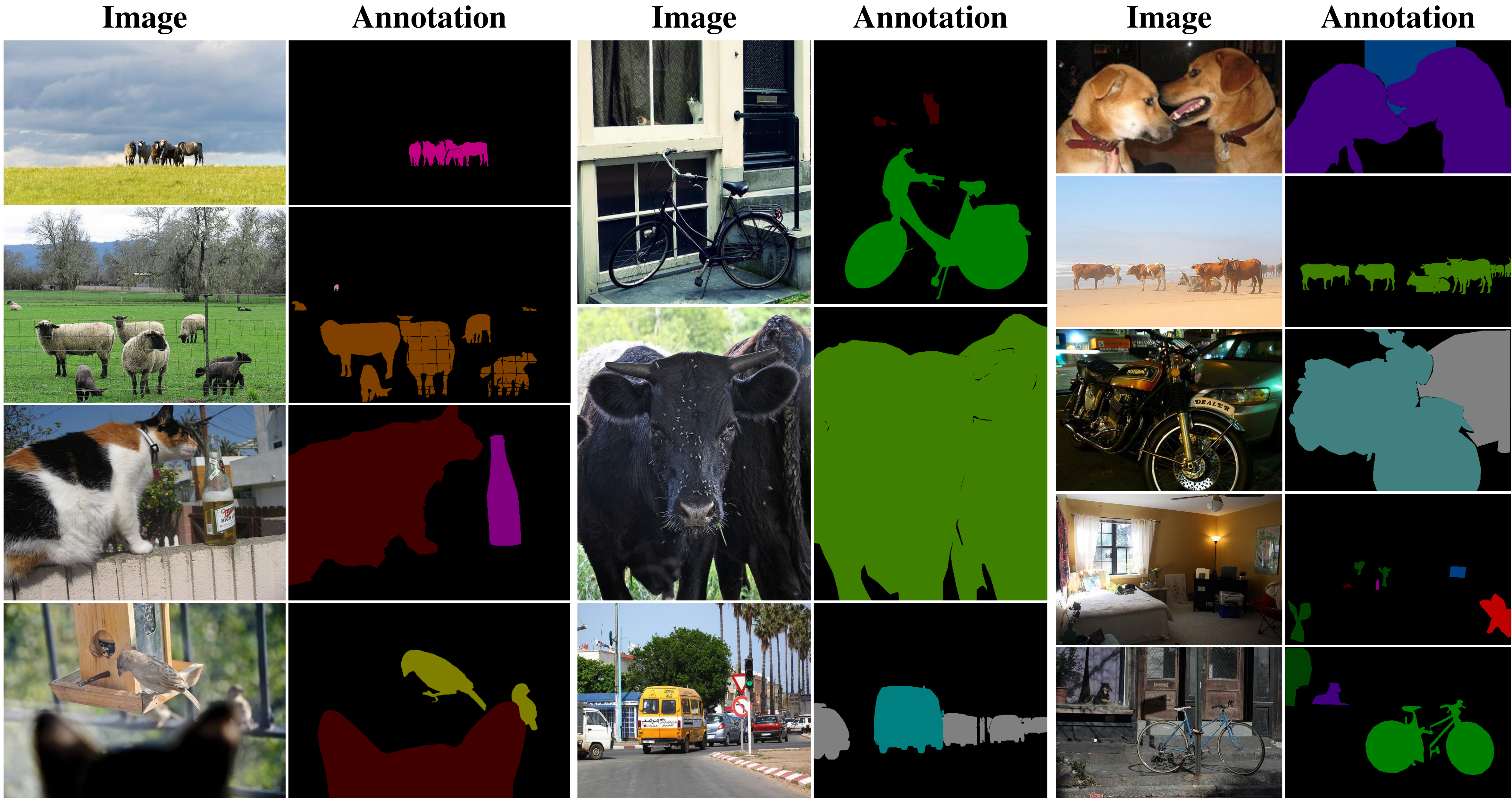}
        \caption{Sample image of PASCAL-B.}
    \label{fig:our_sample}
\end{figure*}

Firstly, we collected images from the LVIS~\cite{lvis} which includes at least one of 20 categories of the PASCAL VOC classes. However, since \texttt{potted plant} class does not exist in the LVIS dataset, we collected images with \texttt{potted plant} class from MS COCO~\cite{COCO}. Then, we converted the annotations which do not belong to the 20 categories of the PASCAL VOC dataset into background class. After finishing the above process, 35,242 images remain. Among the remaining images, a few images have improper annotation as shown in~\fref{fig:our_wrong}. Therefore, two computer vision experts (authors of this paper) manually filtered out such images for two weeks and we had 15,263 images left. Finally, we randomly sampled images to ensure the balance over classes and object size distribution and constructed PASCAL-B which consists of 1,137 images with 20 classes. We give some sample images for the PASCAL-B dataset in~\fref{fig:our_sample}.

\section{Description for evaluated methods}\label{sup_sec:evaluated method}
We choose several methods with different weak-level supervision to validate the comprehensiveness of our metric and method.

\noindent\textbf{Bounding box supervision: BBAM and BANA}
BBAM~\cite{bbam} utilizes the existing object detector Faster R-CNN~\cite{ren2015faster} to highlight the regions where the detector concentrate on. They call these highlighted maps a bounding box attribution map. Then, they expand their bounding box attribution map by introducing a perturbation method. It distinguishes a small subset of the input image that leads to the same prediction as to the original image. Using perturbation methods, they try to diminish the useless information (i.e., background) for the detector.

In BANA~\cite{bana}, Oh et al. find that the background regions around the bounding box are consistent. Based on the observation, they effectively distinguish the foreground and background regions in a bounding box by computing the cosine similarity between features in the bounding box and out of it. Additionally, they try to reduce the effect of noisy labels by utilizing the distances between CNN features and classifier weights.

\noindent\textbf{Saliency supervision: EDAM, NS-ROM and RCA}
\sloppy EDAM~\cite{edam} separates the class-specific information from the whole activation map by applying L2-normalization along the channel dimension. Then it utilizes a self-attention mechanism to highlight similar regions among the series of class-specific activation maps. In the end, it enhances the results by using refined saliency maps with the threshold according to the value of the activation map.

NS-ROM~\cite{nsrom} exploits the objects in non-salient regions. Therefore, they introduce a graph-based global reasoning unit to make the model learn global relations. Also, they filter out the background regions using saliency supervision, while capturing the objects outside the saliency map using class activation maps (CAMs). Finally, they enrich their pseudo masks by setting more ignore pixels to generate new pseudo masks after training the segmentation network. Then they train another segmentation network using new pseudo masks.

RCA~\cite{RCA} bridges the gap between image-level semantic information and pixel-level object regions by regional semantic contrast and aggregation. Regional semantic contrast leverages a memory bank to enforce the embedding of the pseudo region to get close to memory embedding of the same category while pushing away from other categories. Also, they utilize a non-parametric attention module called semantic aggregation. It aggregates memory representations for each image and mines inter-image context to capture more informative dataset-level semantics. 

\noindent\textbf{Natural Language Supervision: CLIM}
CLIM~\cite{Xie_2022_CVPR} is built upon Contrastive Language-Image Pre-training (CLIP)~\cite{Radford2021LearningTV}. Firstly, it additionally defines background classes for each image. Then, CLIM utilizes initial CAM to generate foreground masked-out image $I_{F}$ and background masked-out image $I_{B}$. Lastly, using CLIP, it calculates the cosine similarity between these images and corresponding text category labels. For $I_{F}$, the similarity with a ground-truth label is maximized to gradually expand the activations for the whole foreground objects, while the similarity with the corresponding background label is minimized to decouple the foreground from the background. On the other hand, for $I_{B}$, the similarity with a ground-truth label is minimized to recover more probable foreground contents.

\noindent\textbf{Image supervision: IRN, CDA, AMN and RIB}
IRN~\cite{irn} predicts a displacement of each pixel pointing to the centroid to get the class agnostic map based on the rough semantic segmentation map from CAMs. By incorporating CAMs with a class-agnostic map, it obtains instance-wise CAMs and refines the prediction map by the random-walk algorithm.

CDA~\cite{cda} is proposed to tackle the co-occurrence context information problem for WSSS. It first cuts some simple object instances using predicted segmentation masks by the trained network. Then it augments original images by pasting the obtained instances, and re-train the network with those augmented images.

The authors of AMN~\cite{AMN} raise an issue that global thresholding for CAM can lead to low-quality pseudo mask. To address this problem, they introduce new training objectives which apply per-pixel classification and label conditioning. Per-pixel classification makes discriminative part be reduced while expanding the non-discriminative part. Additionally, label conditioning is used to decrease the activation of non-target classes.

In RIB~\cite{rib}, Lee et al. argue that CAMs focus on the discriminative part because of the information bottleneck problem. The information bottleneck problem is that the only information highly related to tasks remains when the information goes backward of a layer in the network. According to the other works related to information bottleneck theory, it becomes worse with double-sided saturating activation functions such as softmax. Inspired by this, they propose to fine-tune the model with a one-sided saturating function to alleviate information bottleneck while expanding CAMs with global non-discriminative region pooling.

\section{Implementation detail} \label{sup_sec:implementation detail}
All the experiment results of baseline methods~\cite{irn,rib,bbam,cda,bana,AMN,edam,Xie_2022_CVPR,nsrom,RCA} are reproduced by the official code, and we strictly follow the hyper-parameter settings provided by each paper. For the MS COCO dataset, we refer to the settings of RIB~\cite{rib}. 
We set $\tau=5$ for $L_{sw}$ and $\lambda=500$ for $L_{sb}$ in all cases.
For balanced training with our loss function, we train the segmentation networks for 30k iterations for the PASCAL VOC dataset. We use pixel-wise cross-entropy loss for the first 20k, 15k, and 25k iterations, then fine-tune them with $L_{sb}$ until the end of training models for BANA~\cite{bana}, EDAM~\cite{edam}, and others~\cite{irn,nsrom,cda,RCA,Xie_2022_CVPR,AMN}, respectively. For the MS COCO dataset, the number of training iterations is 100k. We train the segmentation network with pixel-wise cross-entropy loss for the first 40K iterations, then fine-tune the network with $L_{sb}$ for the remaining iterations. Note that we do not change all the other hyper-parameters of each baseline model.

All the experiments were done by one GeForce RTX 3090 GPU for PASCAL VOC and two RTX 3090 GPUs for MS COCO, which take 11 hours and 53 hours, respectively.
\section{Elastic weight consolidation} \label{sup_sec:ewc}
Elastic Weight Consolidation (EWC)~\cite{ewc} is a technique for continual learning problem which tries to make the model learn various tasks. EWC aims to find the optimal point for the model to be optimized with several tasks. To achieve this goal, EWC constrains the parameters of the model which have a high correlation with the past training data. In other words, EWC suppresses the change of parameters based on its importance for the previous task. The loss function for EWC is defined as:
 \begin{gather} \label{eq1}
     L_{total}=L_{now}+\sum_{i}\frac{\lambda}{2}F_{i}(\theta_{i}-\theta_{A,i}^{*})^2,
 \end{gather}
where $\lambda$ controls the importance of the previous task. It means that as the value of $\lambda$ gets larger, it suppresses the updates of parameters more. $\text{F}_{i}$ shows the importance of $i$-th parameter for the previous task. It indicates the correlation of parameters with past training data. In~\cite{ewc}, it utilizes the diagonal elements of the Fisher information matrix. Lastly, $(\theta_{i}-\theta_{A,i}^{*})$ is the change of parameter between present model ($i.e.,~\theta_{i}$) and previous model ($i.e.,~\theta_{A,i}^{*}$).

\section{Extension to fully-supervised methods} \label{sup_sec:fully}
In main paper, we demonstrate our evaluation metric, dataset, and loss function for weakly-supervised methods. However, they also can be applied in a fully-supervised manner.~\tref{tab:fully_table} reports the accuracy of fully-supervised methods~\cite{FCN,deeplabv1,DeepLab,pspnet,segmenter} in terms of \texttt{mIoU} and \texttt{IA-mIoU}. It shows the same tendency as the experiment results of weakly-supervised methods except that the performances are generally more increased than the weakly-supervised methods when using our loss function.

\begin{table}[h!]
  \centering
  \fontsize{8}{10}\selectfont  
  \begin{threeparttable}  
    \begin{tabular}{c|ccc}  
    \toprule
    \hline
    % \cmidrule(lr){2-5} \cmidrule(lr){6-9}\cmidrule(lr){10-13}
    Dataset&\multicolumn{3}{c}{PASCAL VOC}\cr
    \hline
    Method&\texttt{mIoU}&\texttt{IA-mIoU}&$\texttt{IA}_{S}$\cr  
    \hline
    FCN~\cite{FCN}&67.8 (\textcolor[rgb]{0,0.5,0}{+0.8})&59.8 (\textcolor[rgb]{0,0.5,0}{+4.9})&17.1 (\textcolor[rgb]{0,0.5,0}{+7.9})\cr
    \hline
    PSP~\cite{pspnet}&76.7 (\textcolor[rgb]{0,0.5,0}{+0.6})&65.2 (\textcolor[rgb]{0,0.5,0}{+5.4})&22.1 (\textcolor[rgb]{0,0.5,0}{+13.2})\cr
    \hline
    DeepLabV1~\cite{deeplabv1}&76.9 (\textcolor[rgb]{0,0.5,0}{+2.0})&65.6 (\textcolor[rgb]{0,0.5,0}{+6.4})&18.9 (\textcolor[rgb]{0,0.5,0}{+13.8}))\cr
    \hline
    DeepLabV2~\cite{DeepLab}&77.8 (\textcolor[rgb]{0,0.5,0}{+0.6})&65.8 (\textcolor[rgb]{0,0.5,0}{+3.7})&18.8 (\textcolor[rgb]{0,0.5,0}{+5.6})\cr
    \hline
    Segmentor~\cite{segmenter}&79.9 (\textcolor[rgb]{0,0.5,0}{+0.6})&69.5 (\textcolor[rgb]{0,0.5,0}{+5.2})&24.1 (\textcolor[rgb]{0,0.5,0}{+16.7}))\cr
    \toprule
    \bottomrule
    Dataset&\multicolumn{3}{c}{PASCAL B}\cr
    \hline
    Method&\texttt{mIoU}&\texttt{IA-mIoU}&$\texttt{IA}_{S}$\cr  
    \hline
    FCN~\cite{FCN}&56.6 (\textcolor[rgb]{0,0.5,0}{+1.2})&40.3 (\textcolor[rgb]{0,0.5,0}{+5.0})&10.1 (\textcolor[rgb]{0,0.5,0}{+5.5})\cr
    \hline
    PSP~\cite{pspnet}&63.3 (\textcolor[rgb]{0,0.5,0}{+0.1})&42.4 (\textcolor[rgb]{0,0.5,0}{+4.9})&13.4 (\textcolor[rgb]{0,0.5,0}{+6.3})\cr
    \hline
    DeepLabV1~\cite{deeplabv1}&65.7 (\textcolor[rgb]{0,0.5,0}{+1.3})&45.4 (\textcolor[rgb]{0,0.5,0}{+5.8})&13.3 (\textcolor[rgb]{0,0.5,0}{+7.1})\cr
    \hline
    DeepLabV2~\cite{DeepLab}&66.6 (\textcolor[rgb]{0,0.5,0}{+1.3})&46.2 (\textcolor[rgb]{0,0.5,0}{+3.2})&15.6 (\textcolor[rgb]{0,0.5,0}{+4.2})\cr
    \hline
    Segmentor~\cite{segmenter}&67.9 (\textcolor[rgb]{1,0,0}{$-0.2$})&45.9 (\textcolor[rgb]{0,0.5,0}{+4.9})&13.1 (\textcolor[rgb]{0,0.5,0}{+7.6})\cr
    \hline
    \bottomrule  
    \end{tabular}  
    \end{threeparttable}
  \caption{Experimental results of fully-supervised method for PASCAL VOC and PASCAL-B.}
    \label{tab:fully_table}  
\end{table}

\section{Qualitative result} \label{sup_sec:qualitative result}
\begin{figure*}[h!]
    \centering
        \includegraphics[width=\linewidth]{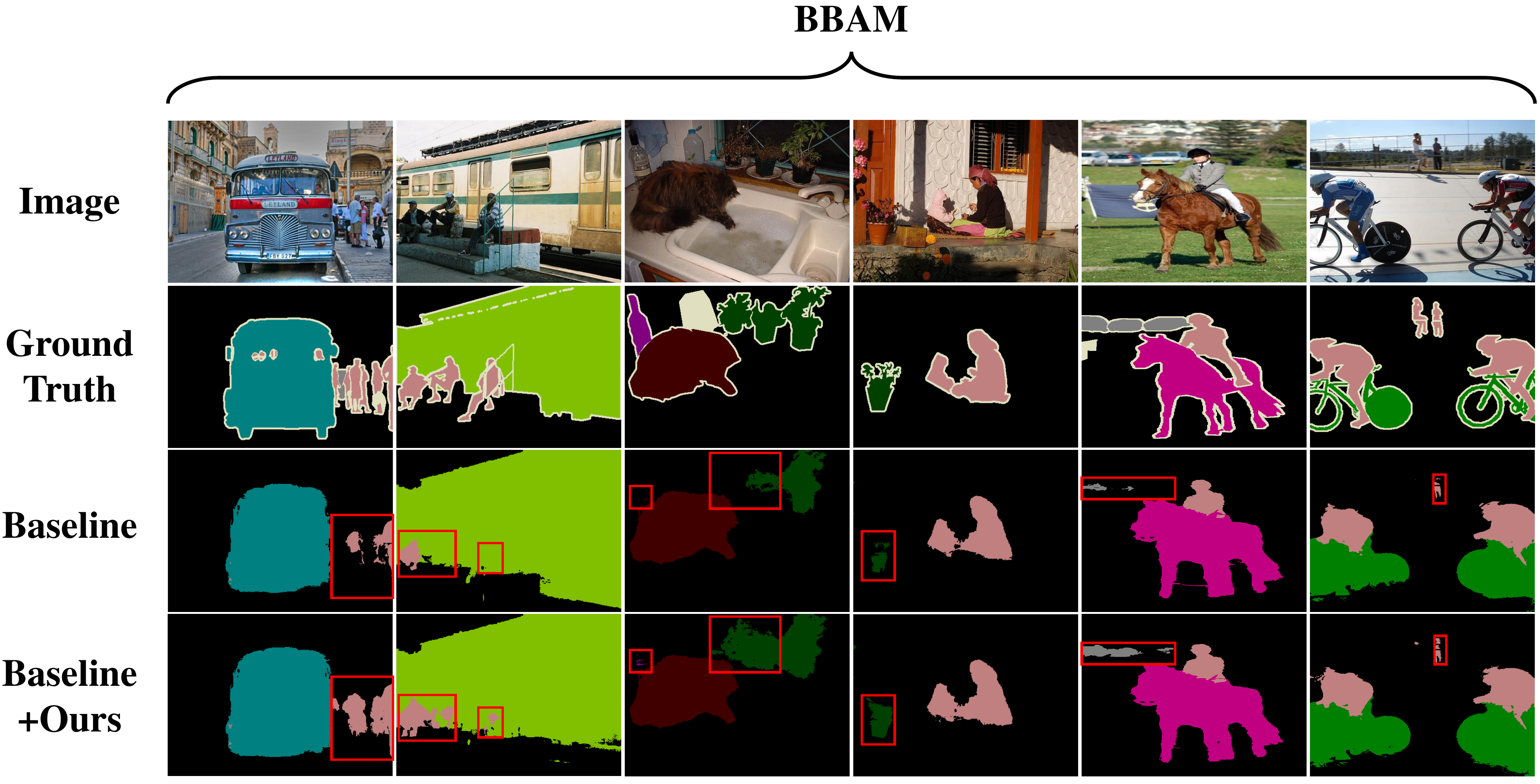}
        \caption{Visualization of BBAM on PASCAL VOC. Our loss function successfully fine-tunes baseline model to improve the ability of capturing objects including small-sized ones which is expressed by red bounding boxes.}
    \label{fig:voc_bbam}
\end{figure*}

\begin{figure*}[t!]
    \centering
        \includegraphics[width=\linewidth]{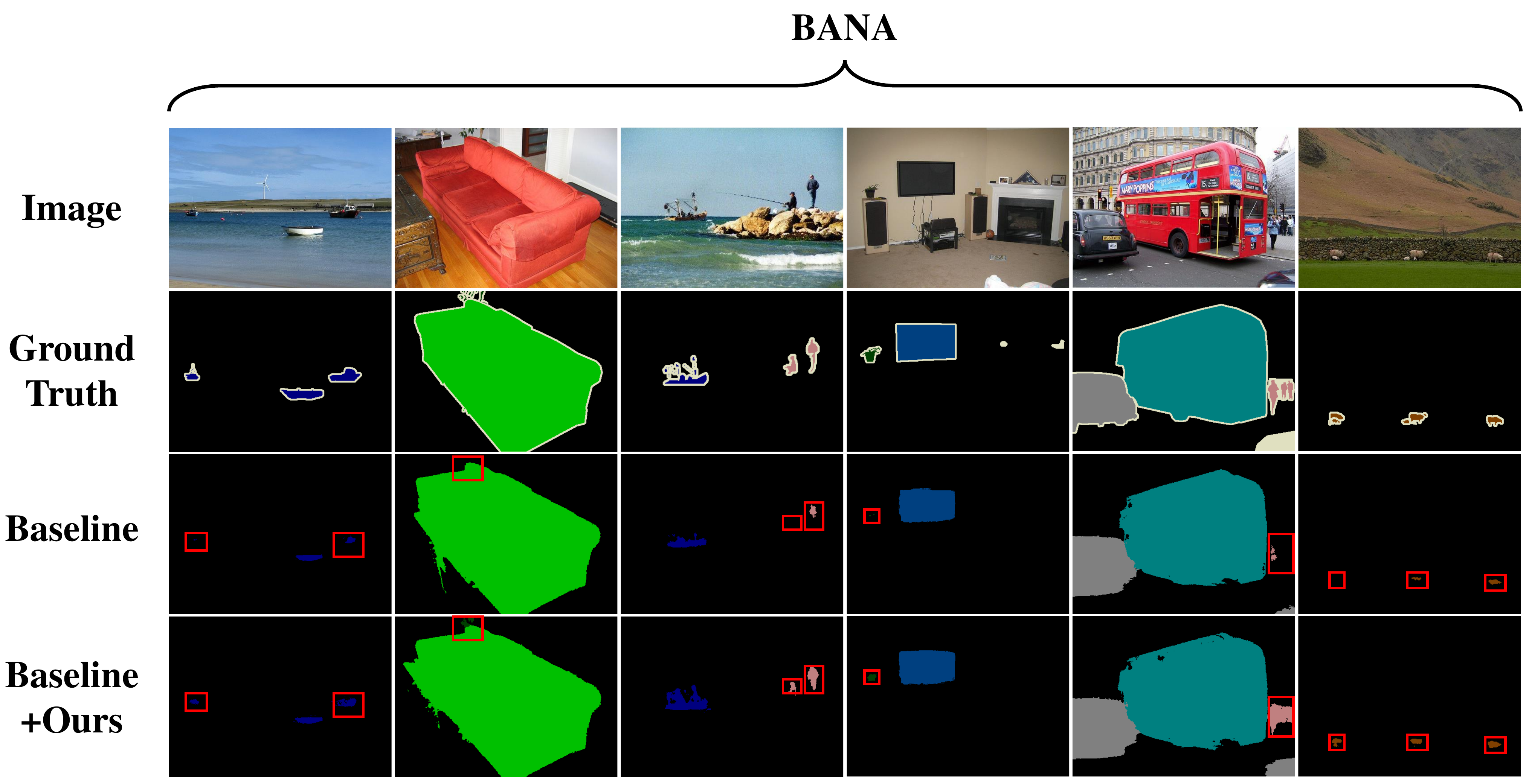}
        \caption{Visualization of BANA on PASCAL VOC. Our loss function successfully fine-tunes baseline model to improve the ability of capturing objects including small-sized ones which is expressed by red bounding boxes.}
    \label{fig:voc_bana}
\end{figure*}

\begin{figure*}[t!]
    \centering
        \includegraphics[width=\linewidth]{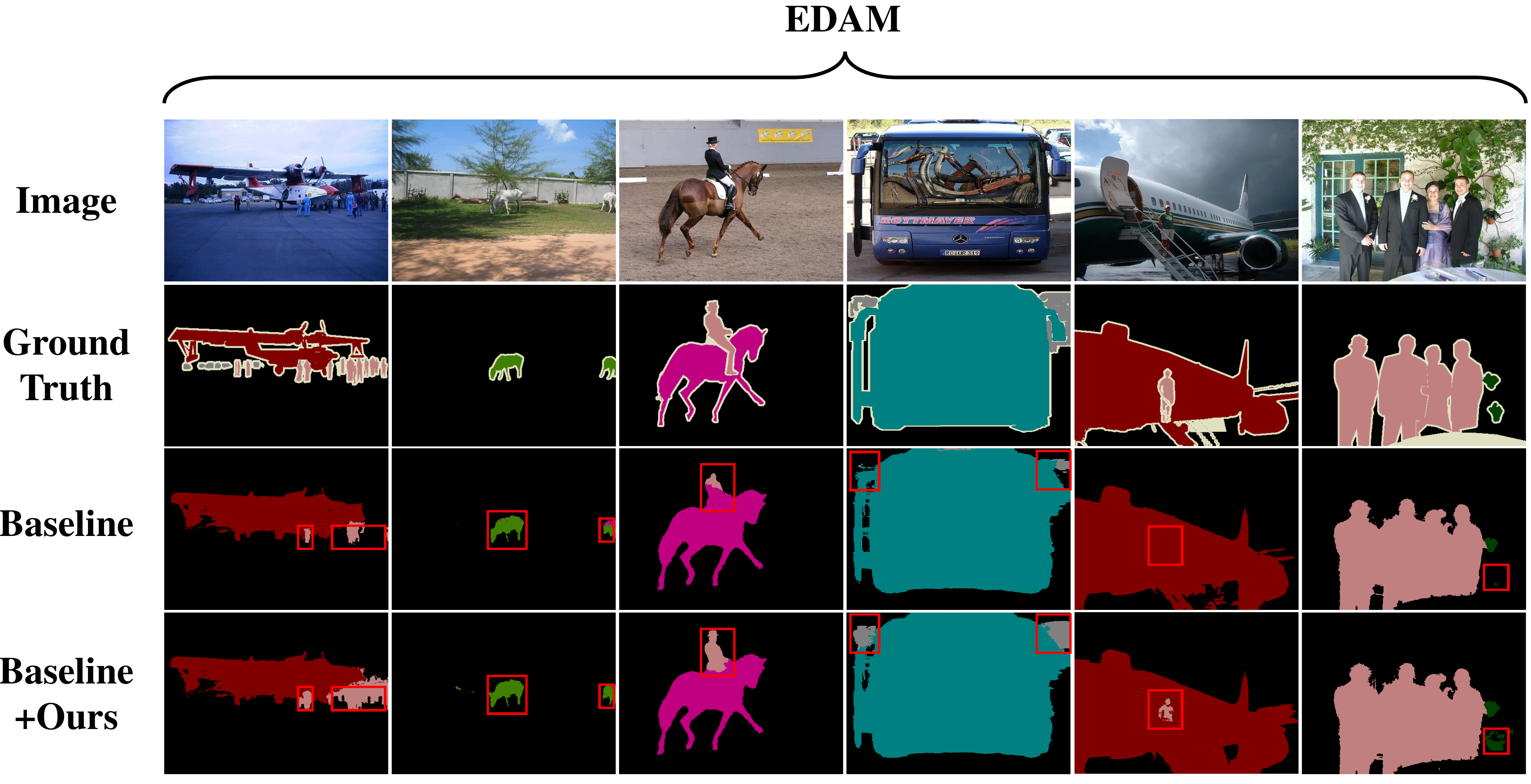}
        \caption{Visualization of EDAM on PASCAL VOC. Our loss function successfully fine-tunes baseline model to improve the ability of capturing objects including small-sized ones which is expressed by red bounding boxes.}
    \label{fig:voc_edam}
\end{figure*}

\begin{figure*}[t!]
    \centering
        \includegraphics[width=\linewidth]{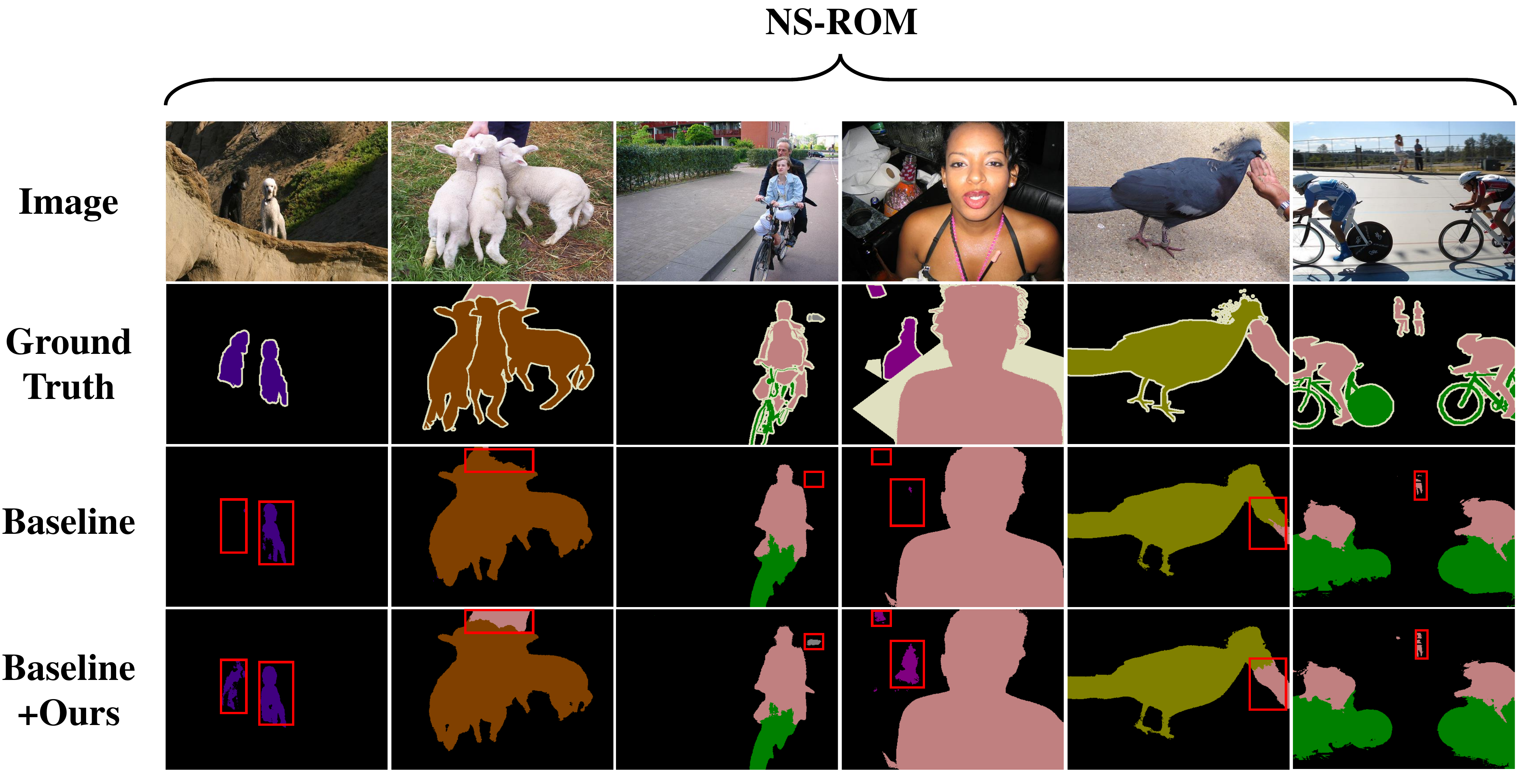}
        \caption{Visualization of NS-ROM on PASCAL VOC. Our loss function successfully fine-tunes baseline model to improve the ability of capturing objects including small-sized ones which is expressed by red bounding boxes.}
    \label{fig:voc_nsrom}
\end{figure*}

\begin{figure*}[t!]
    \centering
        \includegraphics[width=\linewidth]{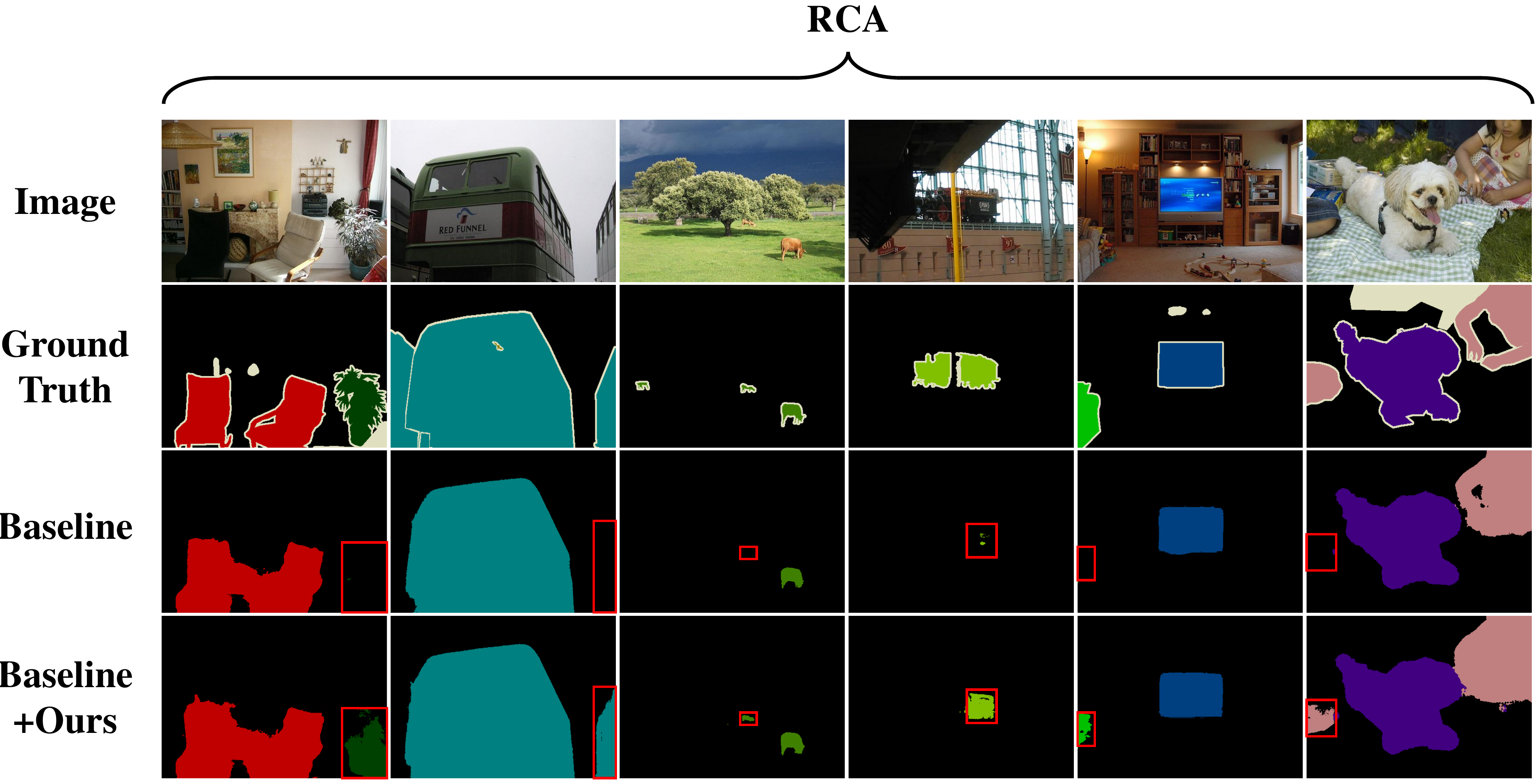}
        \caption{Visualization of RCA on PASCAL VOC. Our loss function successfully fine-tunes baseline model to improve the ability of capturing objects including small-sized ones which is expressed by red bounding boxes.}
    \label{fig:voc_rca}
\end{figure*}

\begin{figure*}[t!]
    \centering
        \includegraphics[width=\linewidth]{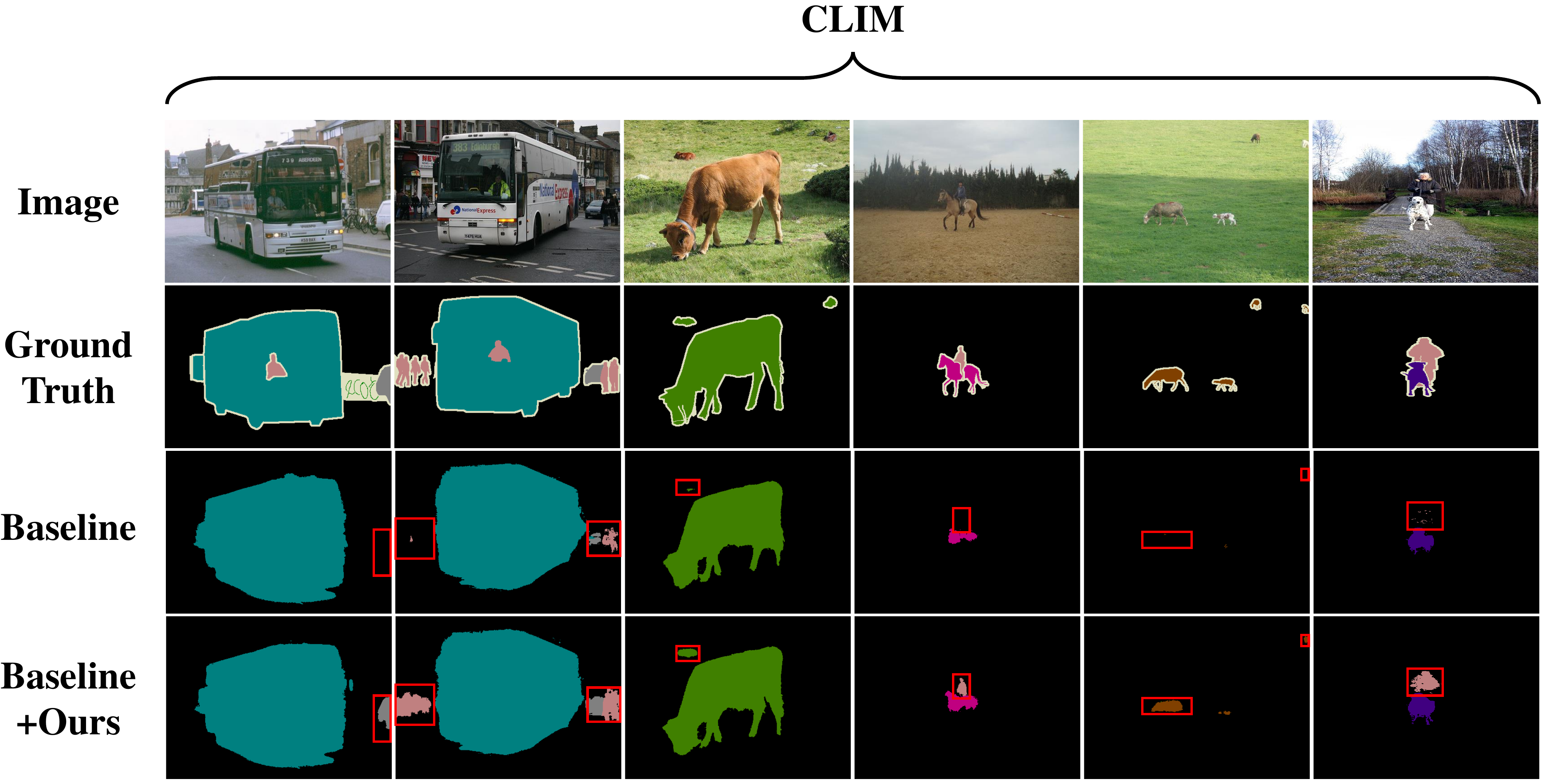}
        \caption{Visualization of CLIM on PASCAL VOC. Our loss function successfully fine-tunes baseline model to improve the ability of capturing objects including small-sized ones which is expressed by red bounding boxes.}
    \label{fig:voc_clim}
\end{figure*}

\begin{figure*}[t!]
    \centering
        \includegraphics[width=\linewidth]{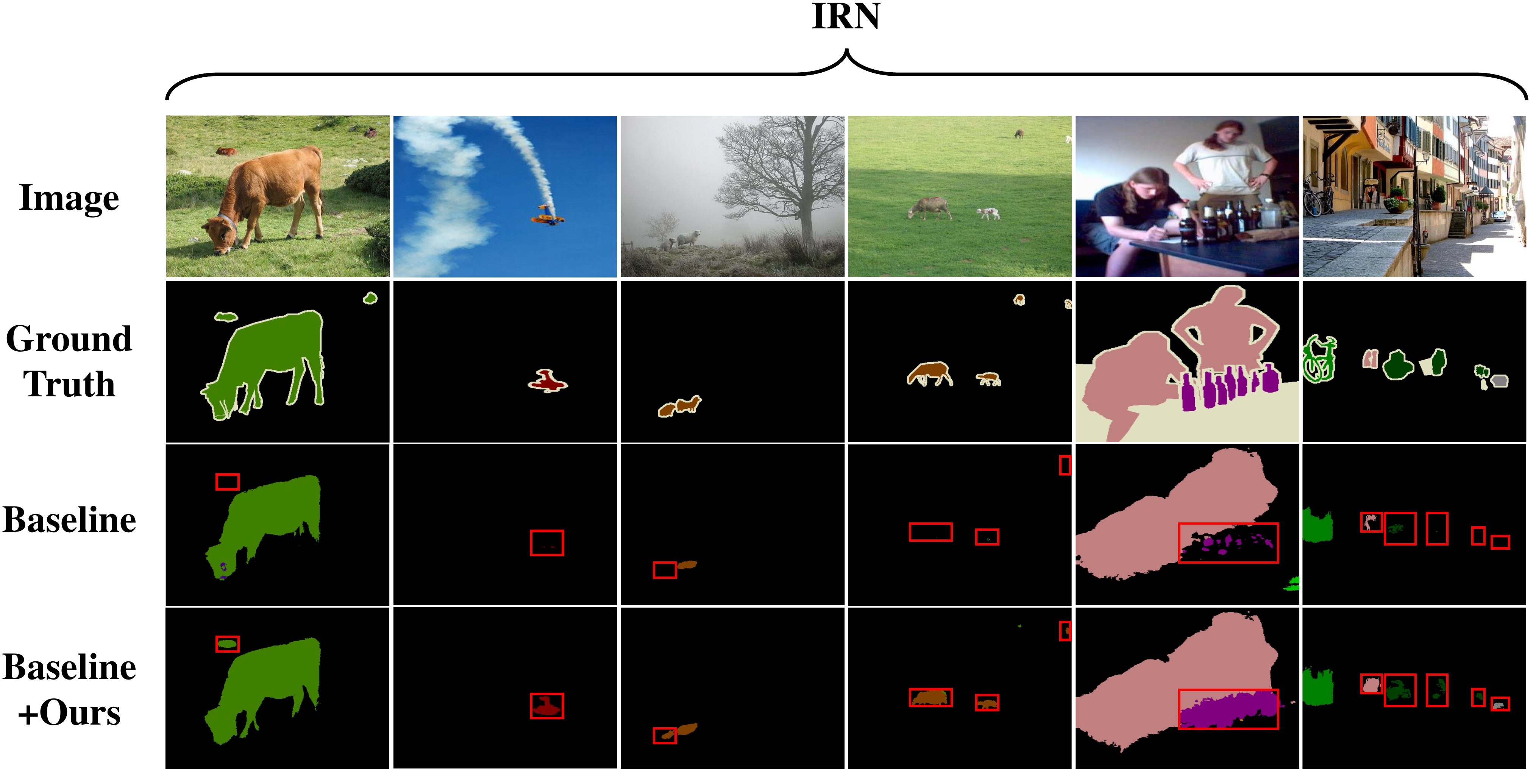}
        \caption{Visualization of IRN on PASCAL VOC. Our loss function successfully fine-tunes baseline model to improve the ability of capturing objects including small-sized ones which is expressed by red bounding boxes.}
    \label{fig:voc_irn}
\end{figure*}

\begin{figure*}[t!]
    \centering
        \includegraphics[width=\linewidth]{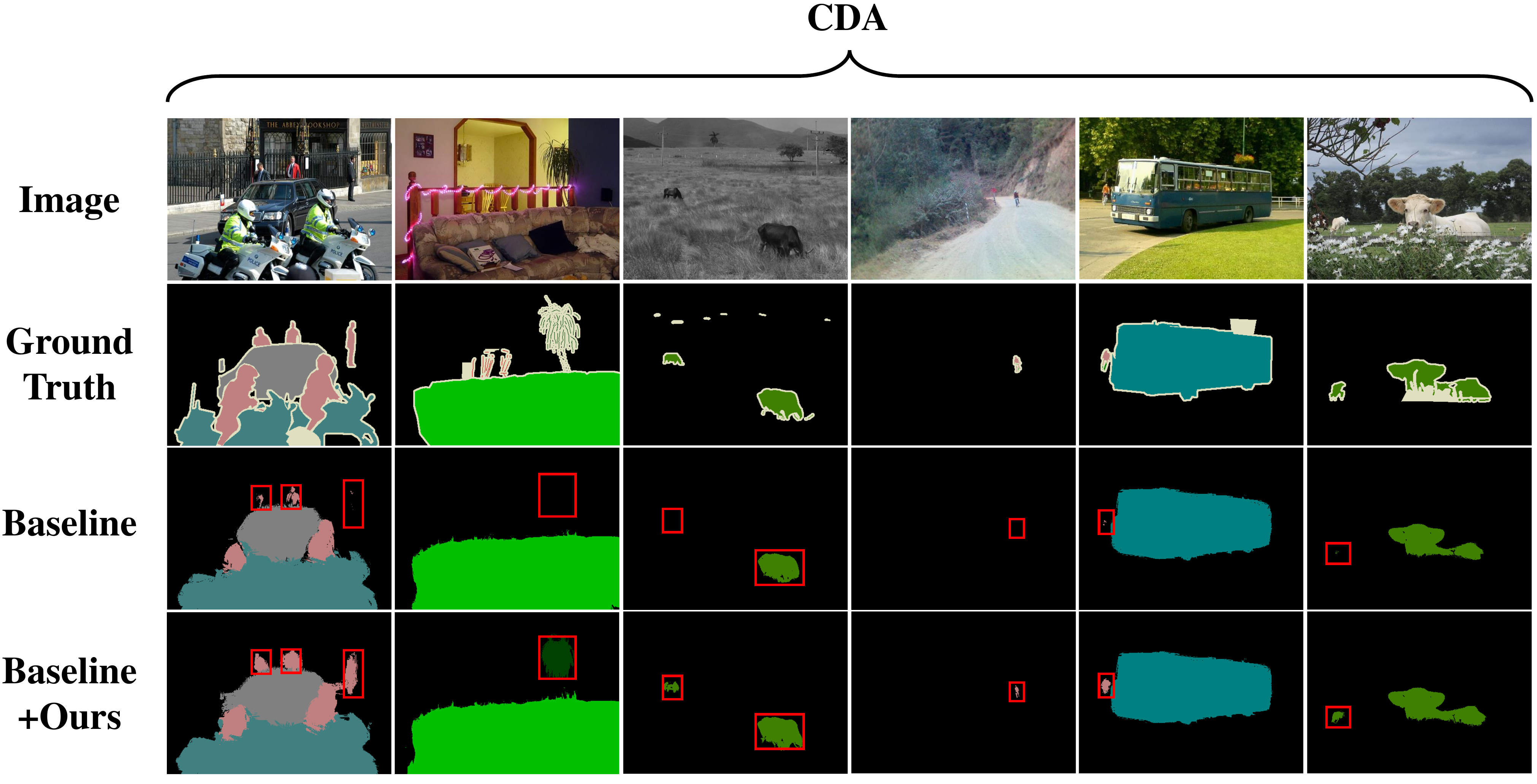}
        \caption{Visualization of CDA on PASCAL VOC. Our loss function successfully fine-tunes baseline model to improve the ability of capturing objects including small-sized ones which is expressed by red bounding boxes.}
    \label{fig:voc_cda}
\end{figure*}

\begin{figure*}[t!]
    \centering
        \includegraphics[width=\linewidth]{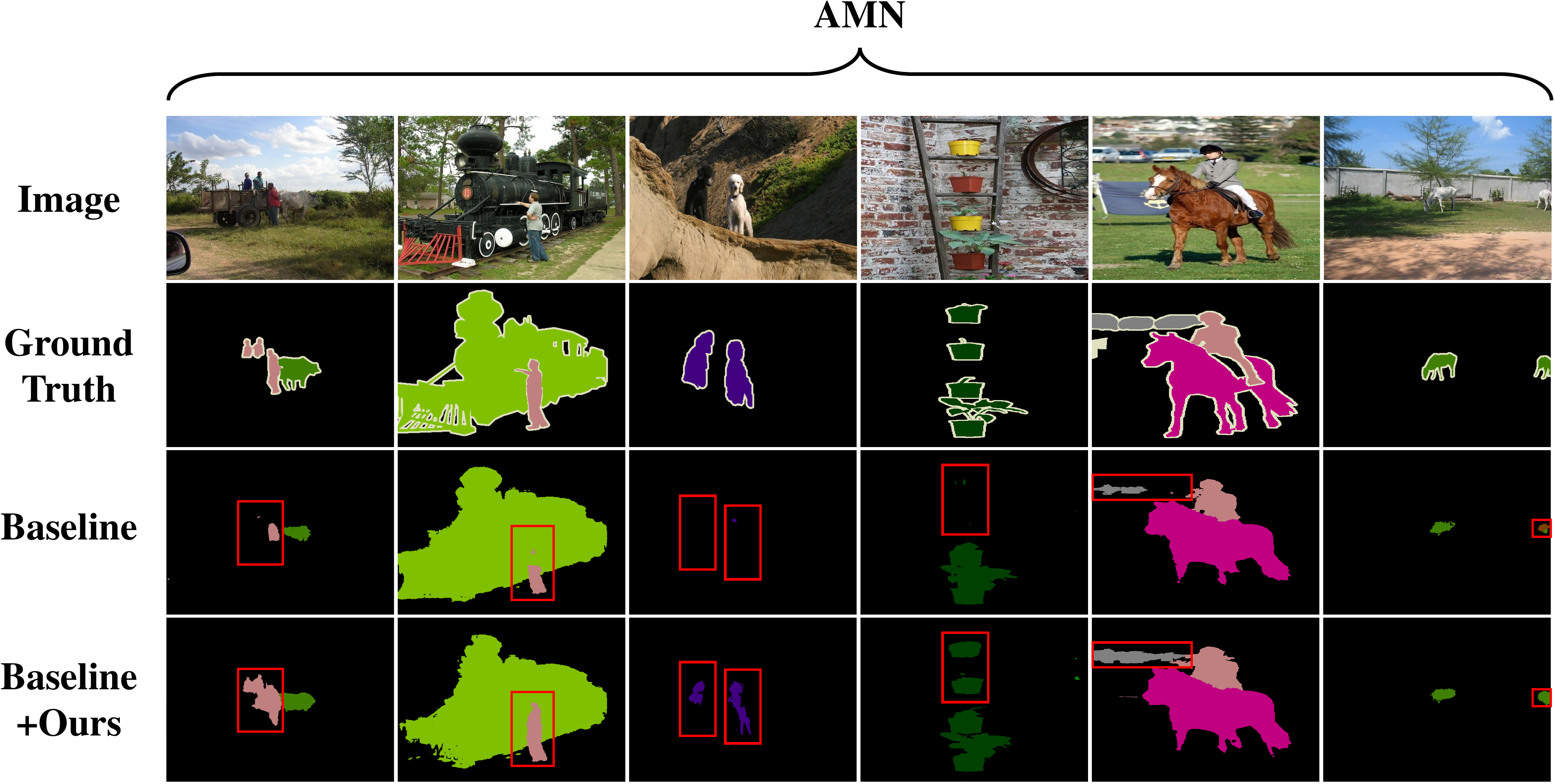}
        \caption{Visualization of AMN on PASCAL VOC. Our loss function successfully fine-tunes baseline model to improve the ability of capturing objects including small-sized ones which is expressed by red bounding boxes.}
    \label{fig:voc_amn}
\end{figure*}

\begin{figure*}[t!]
    \centering
        \includegraphics[width=\linewidth]{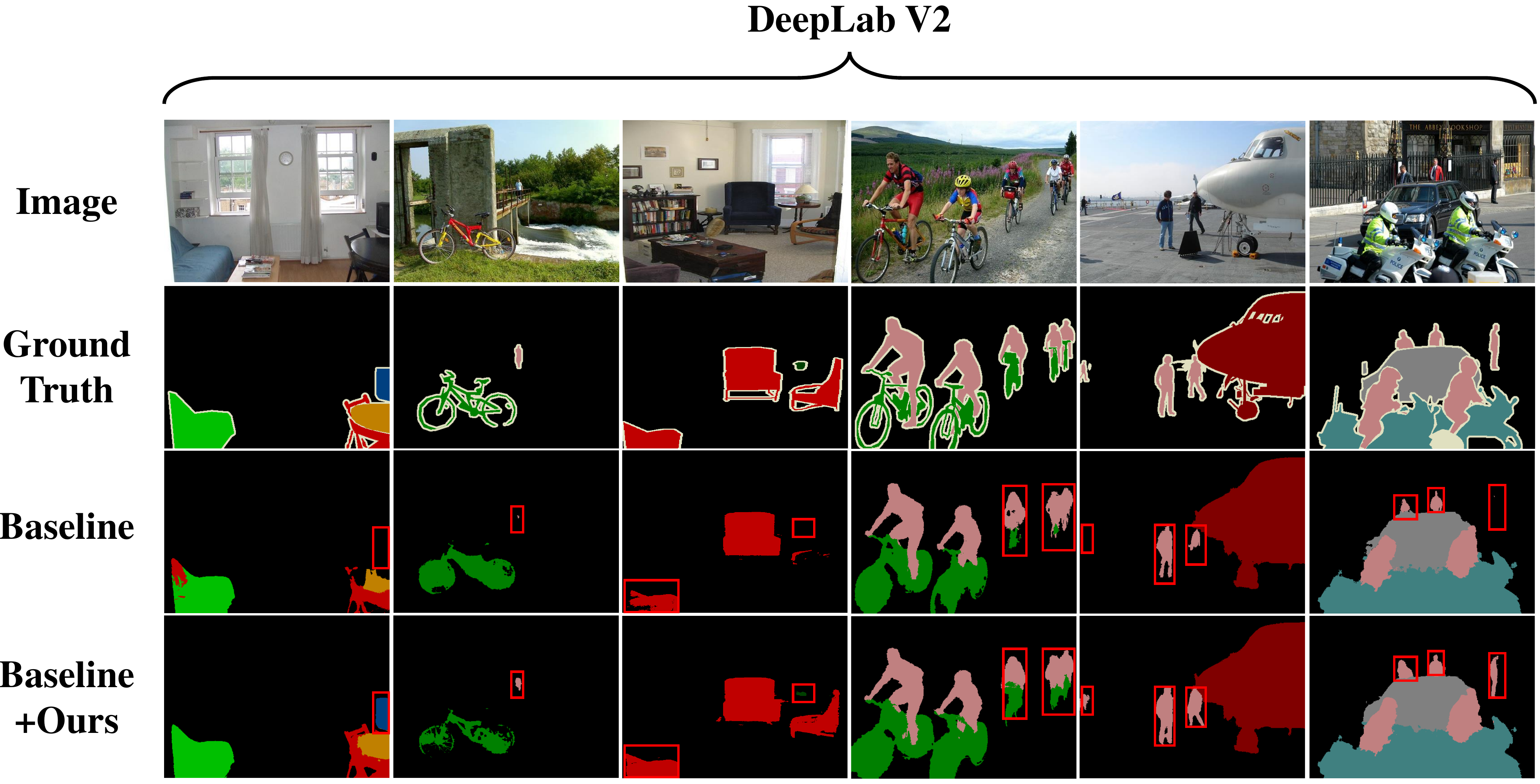}
        \caption{Visualization of DeepLab V2 on PASCAL VOC. Our loss function successfully fine-tunes baseline model to improve the ability of capturing objects including small-sized ones which is expressed by red bounding boxes.}
    \label{fig:voc_deeplab2}
\end{figure*}

\begin{figure*}[t!]
    \centering
        \includegraphics[width=\linewidth]{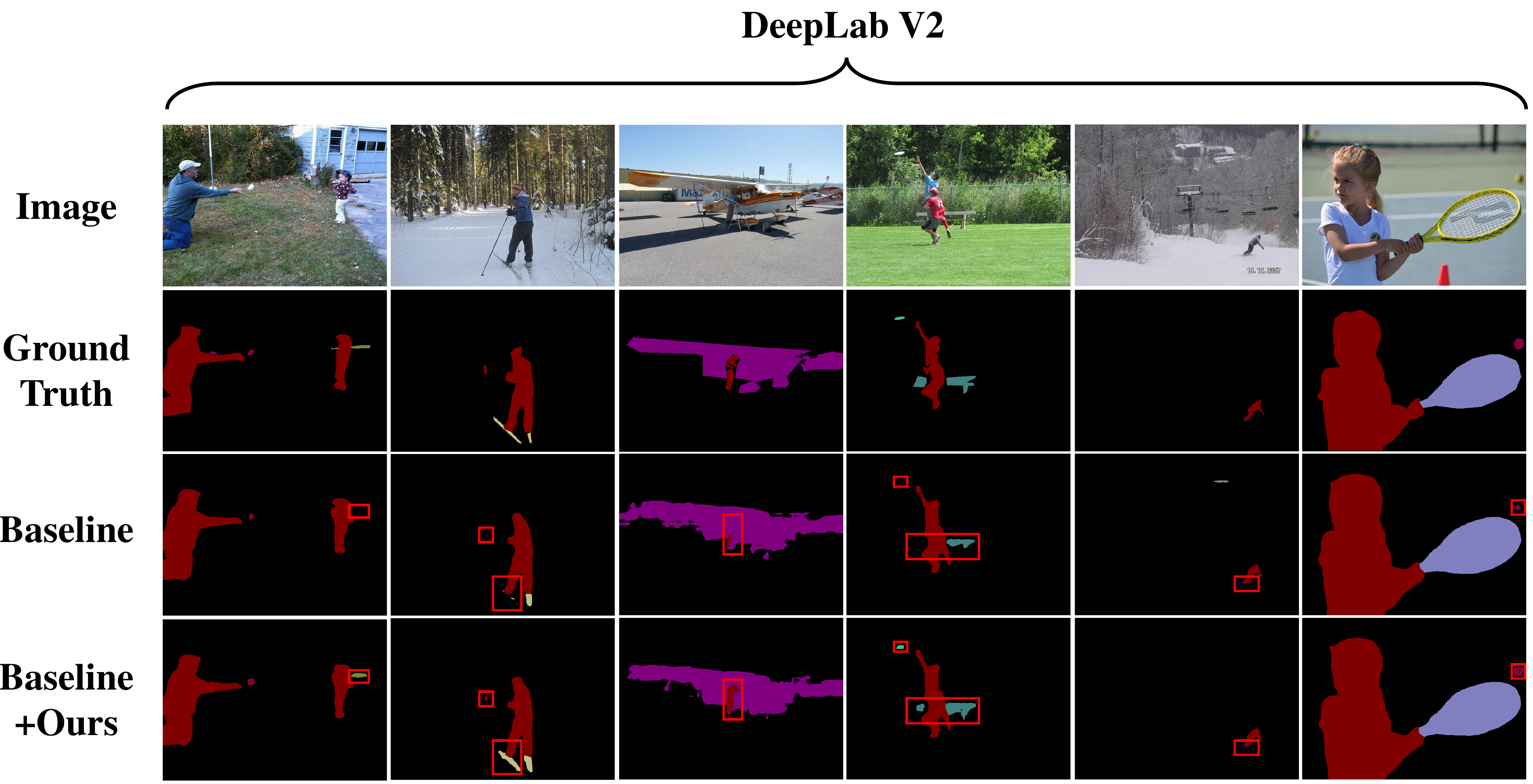}
        \caption{Visualization of DeepLab V2 on MS COCO. Our loss function successfully fine-tunes baseline model to improve the ability of capturing objects including small-sized ones which is expressed by red bounding boxes.}
    \label{fig:coco_fully}
\end{figure*}

\begin{figure*}[t!]
    \centering
        \includegraphics[width=\linewidth]{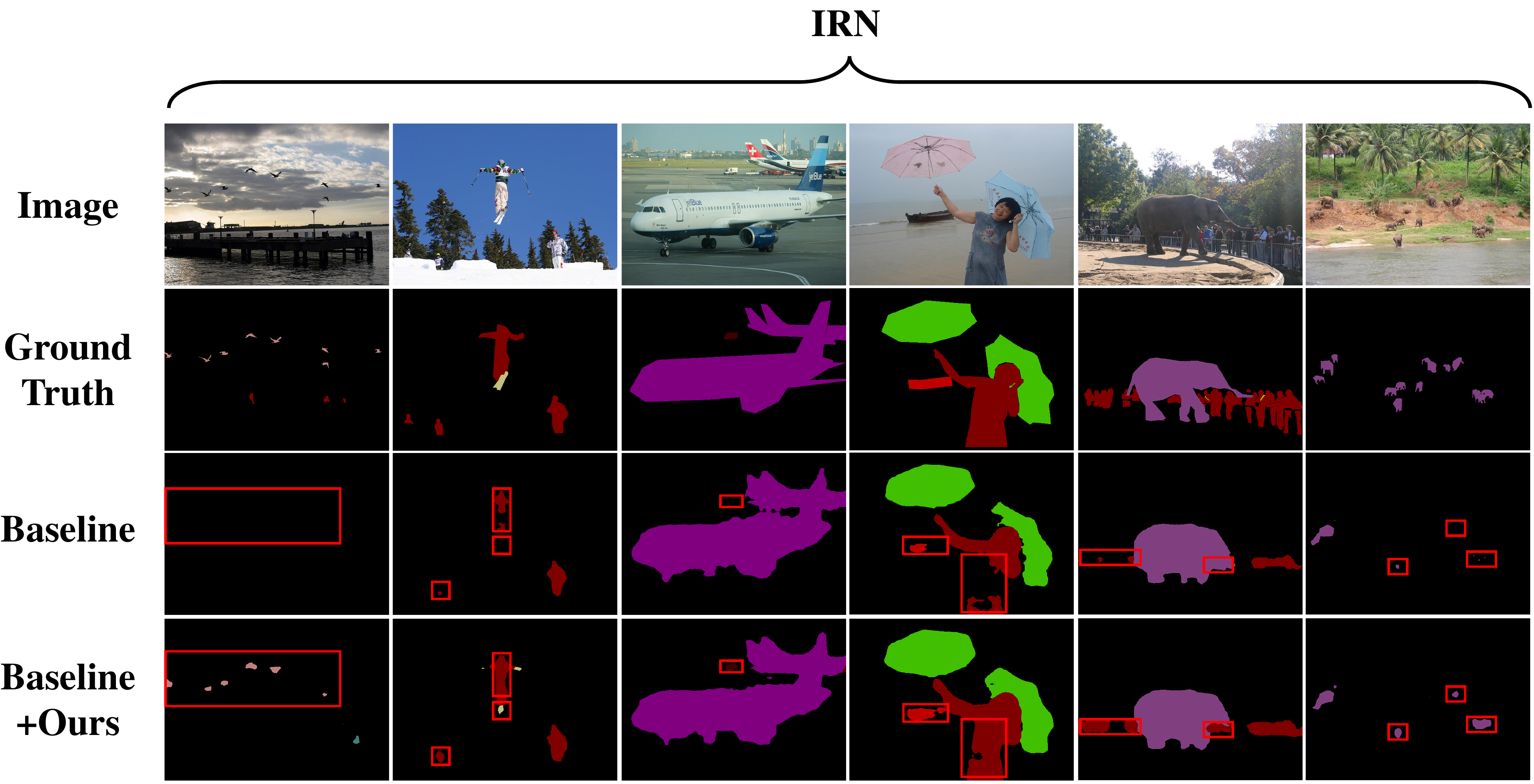}
        \caption{Visualization of IRN on MS COCO. Our loss function successfully fine-tunes baseline model to improve the ability of capturing objects including small-sized ones which is expressed by red bounding boxes.}
    \label{fig:coco_irn}
\end{figure*}

\begin{figure*}[t!]
    \centering
        \includegraphics[width=\linewidth]{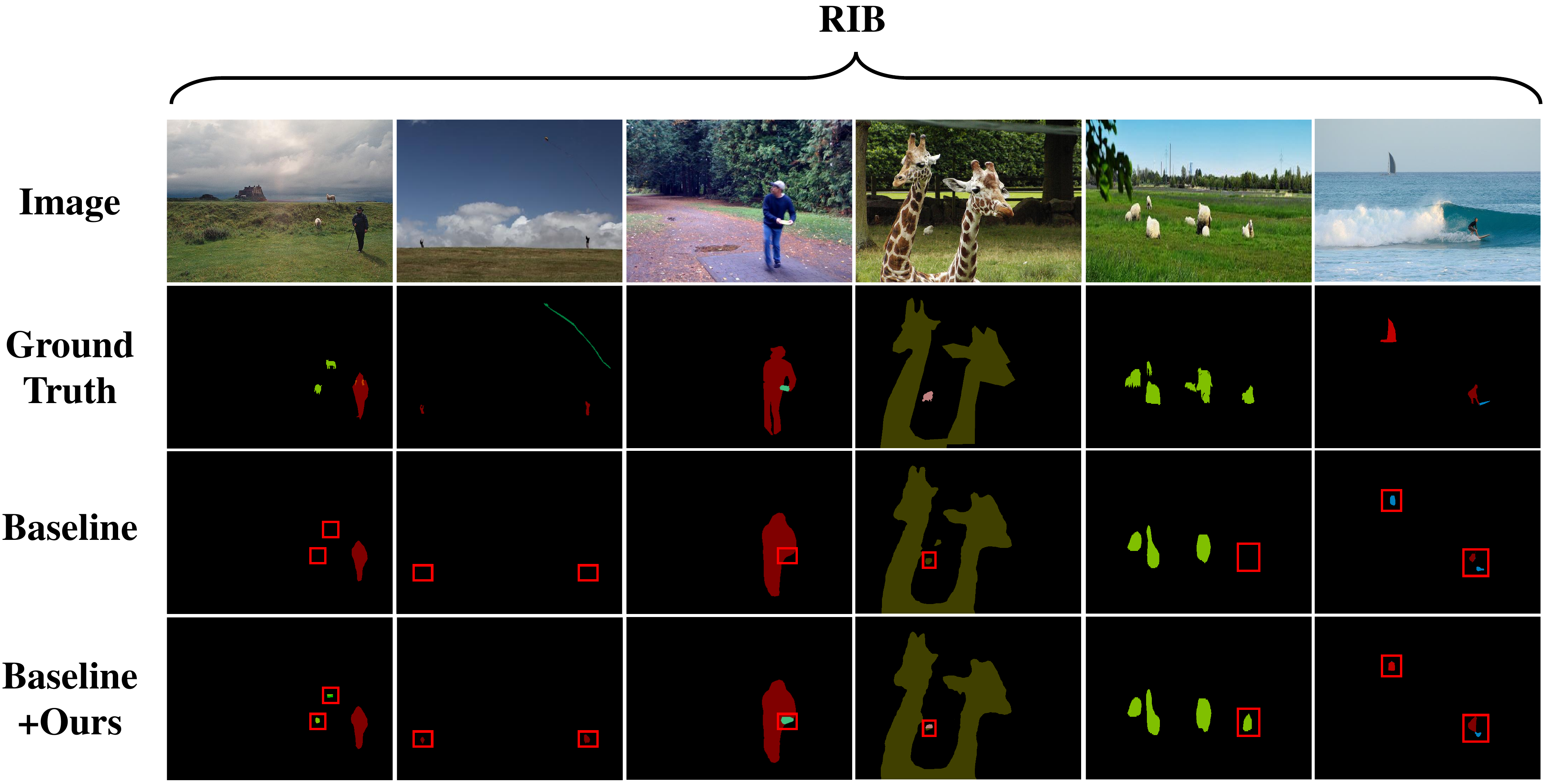}
        \caption{Visualization of RIB on MS COCO. Our loss function successfully fine-tunes baseline model to improve the ability of capturing objects including small-sized ones which is expressed by red bounding boxes.}
    \label{fig:coco_rib}
\end{figure*}

\begin{figure*}[t!]
    \centering
        \includegraphics[width=\linewidth]{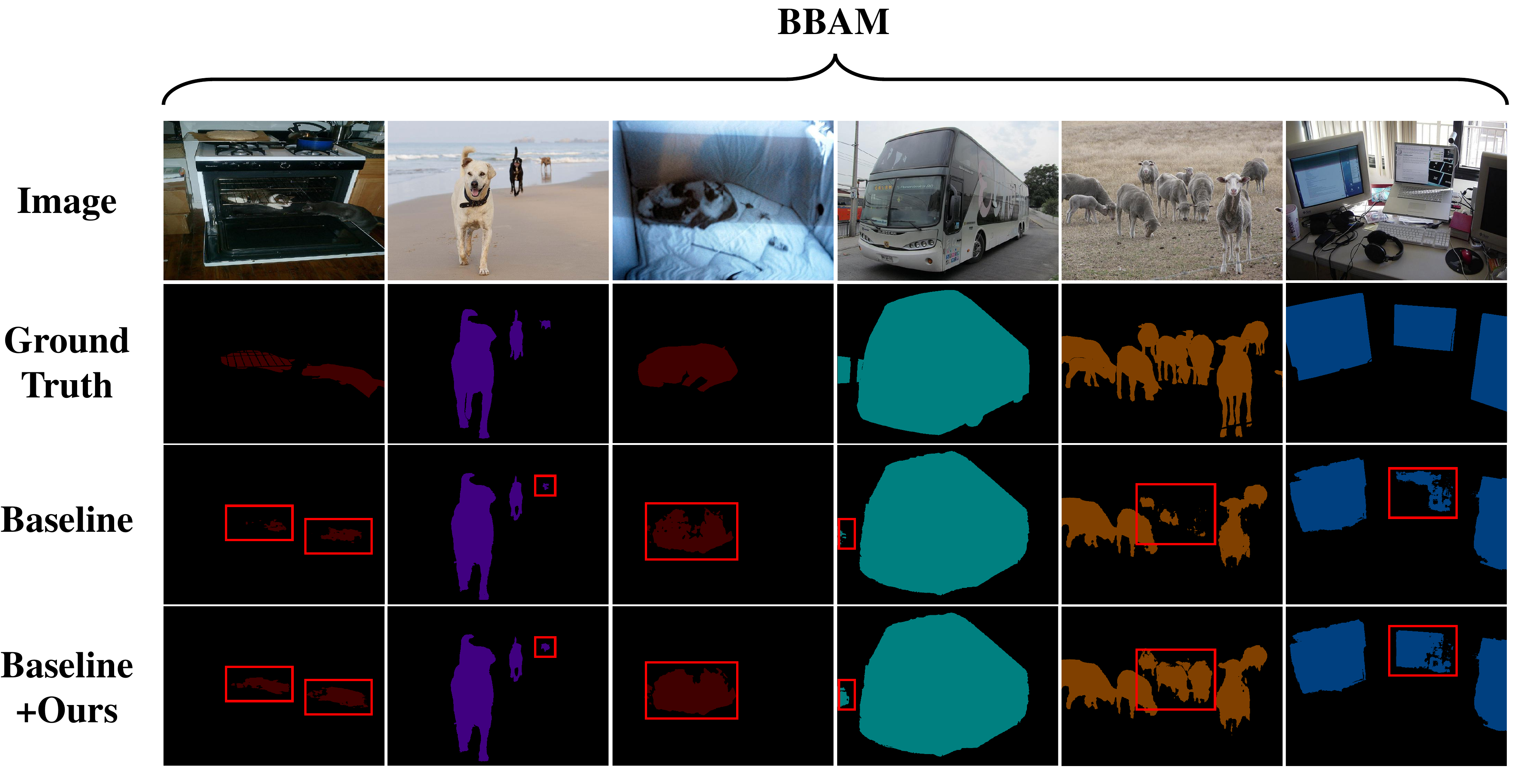}
        \caption{Visualization of BBAM on PASCAL-B. Our loss function successfully fine-tunes baseline model to improve the ability of capturing objects including small-sized ones which is expressed by red bounding boxes.}
    \label{fig:our_bbam}
\end{figure*}

\begin{figure*}[t!]
    \centering
        \includegraphics[width=\linewidth]{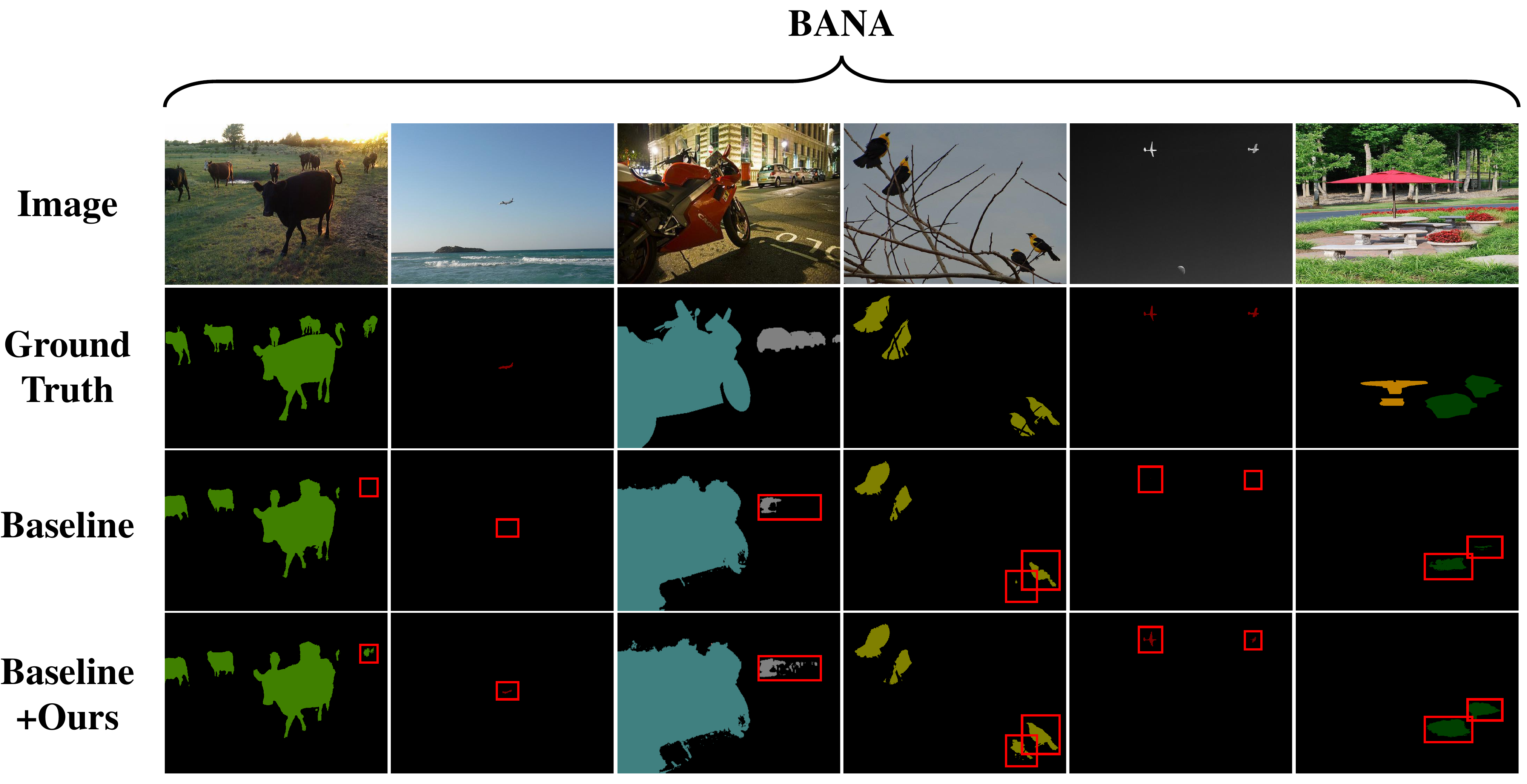}
        \caption{Visualization of BANA on PASCAL-B. Our loss function successfully fine-tunes baseline model to improve the ability of capturing objects including small-sized ones which is expressed by red bounding boxes.}
    \label{fig:our_bana}
\end{figure*}

\begin{figure*}[t!]
    \centering
        \includegraphics[width=\linewidth]{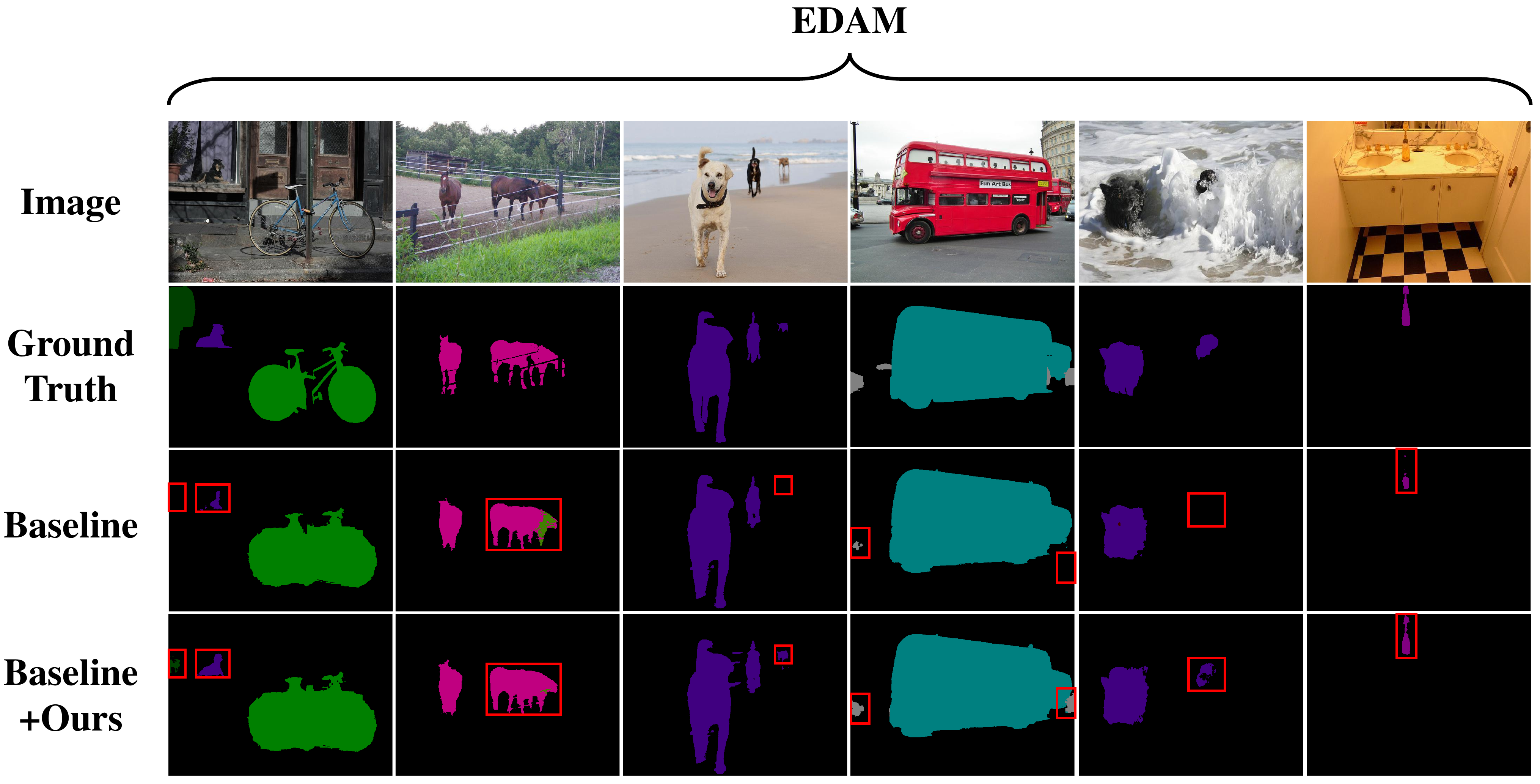}
        \caption{Visualization of EDAM on PASCAL-B. Our loss function successfully fine-tunes baseline model to improve the ability of capturing objects including small-sized ones which is expressed by red bounding boxes.}
    \label{fig:our_edam}
\end{figure*}

\begin{figure*}[t!]
    \centering
        \includegraphics[width=\linewidth]{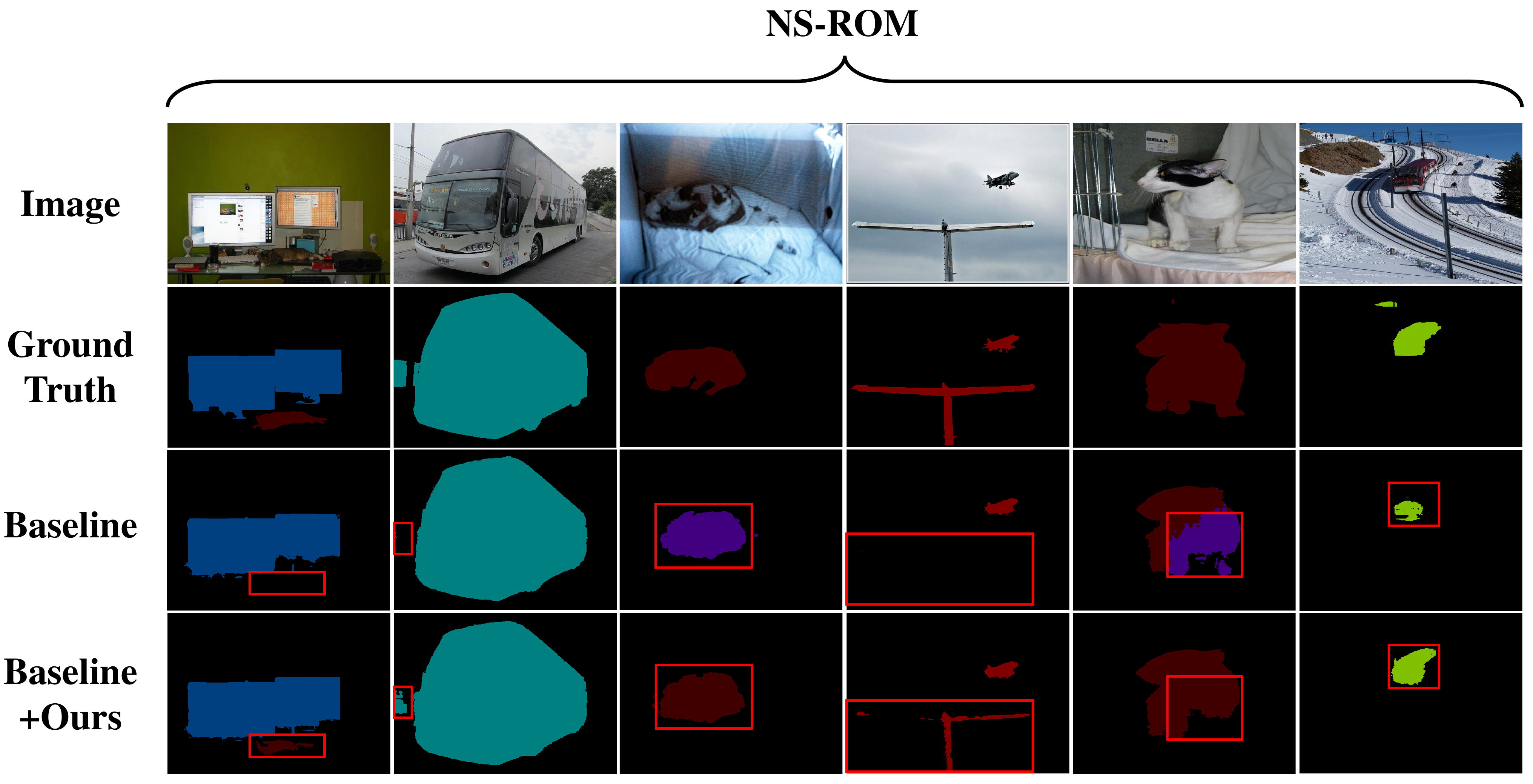}
        \caption{Visualization of NS-ROM on PASCAL-B. Our loss function successfully fine-tunes baseline model to improve the ability of capturing objects including small-sized ones which is expressed by red bounding boxes.}
    \label{fig:our_nsrom}
\end{figure*}

\begin{figure*}[t!]
    \centering
        \includegraphics[width=\linewidth]{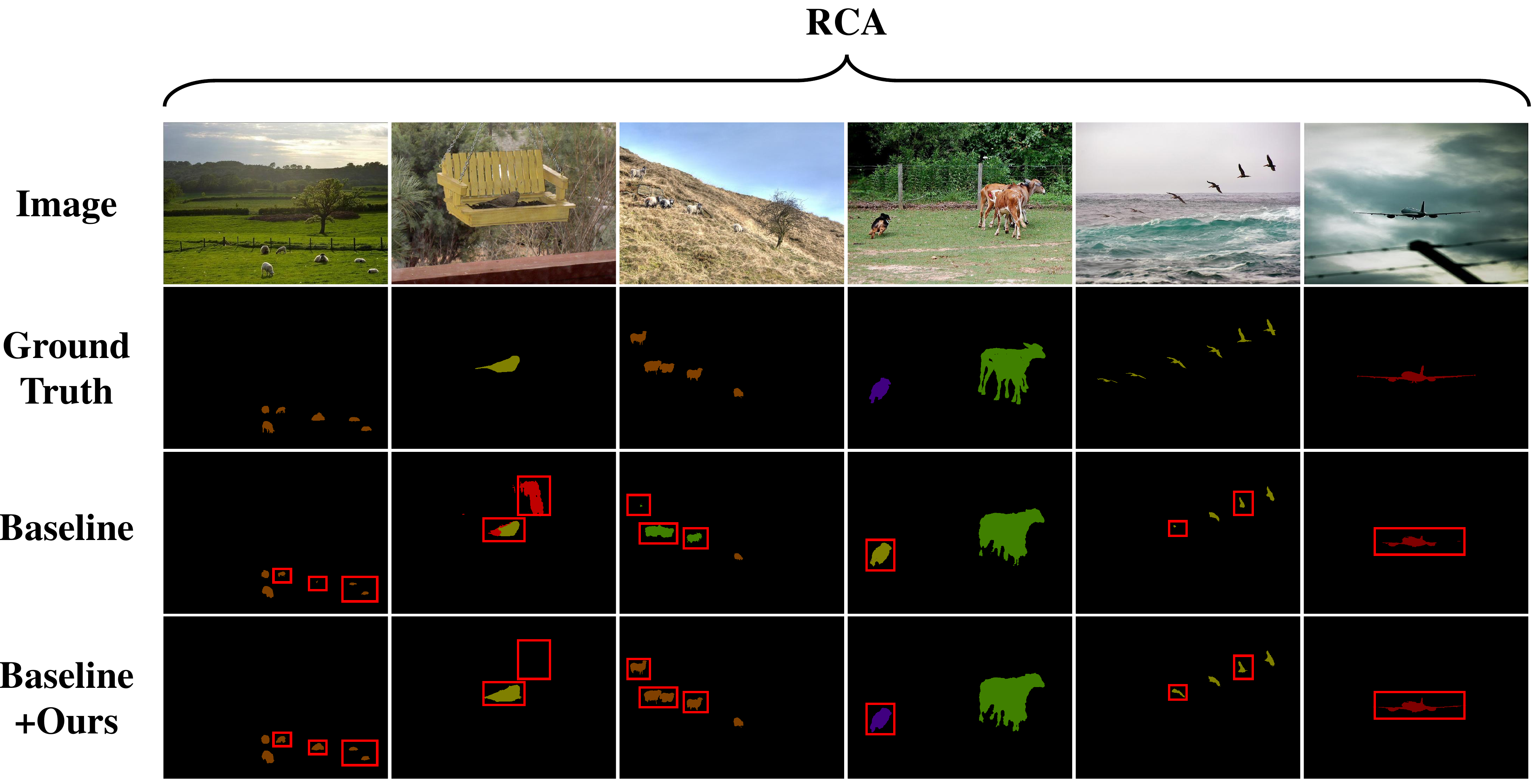}
        \caption{Visualization of RCA on PASCAL-B. Our loss function successfully fine-tunes baseline model to improve the ability of capturing objects including small-sized ones which is expressed by red bounding boxes.}
    \label{fig:our_rca}
\end{figure*}

\begin{figure*}[t!]
    \centering
        \includegraphics[width=\linewidth]{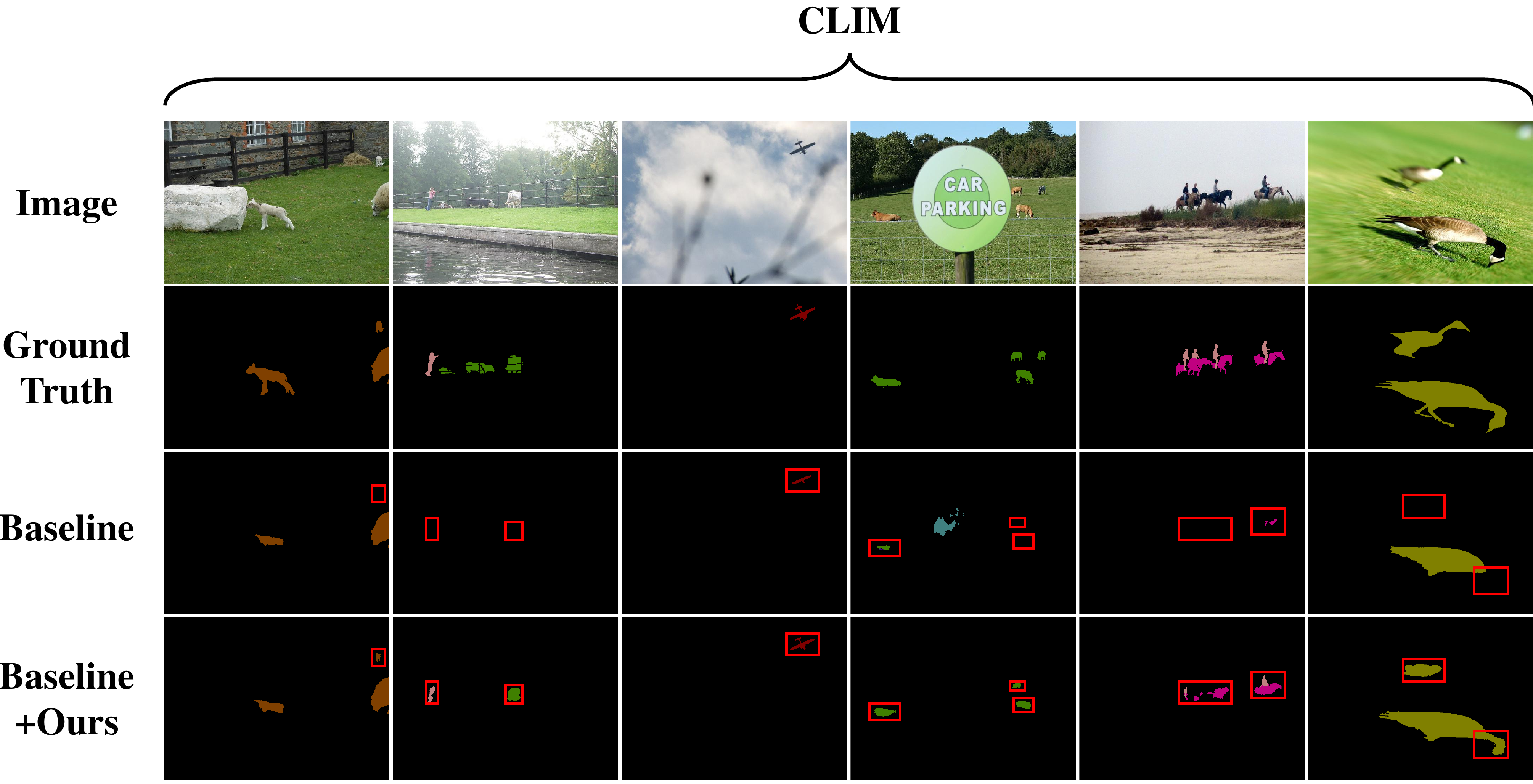}
        \caption{Visualization of CLIM on PASCAL-B. Our loss function successfully fine-tunes baseline model to improve the ability of capturing objects including small-sized ones which is expressed by red bounding boxes.}
    \label{fig:our_clim}
\end{figure*}

\begin{figure*}[t!]
    \centering
        \includegraphics[width=\linewidth]{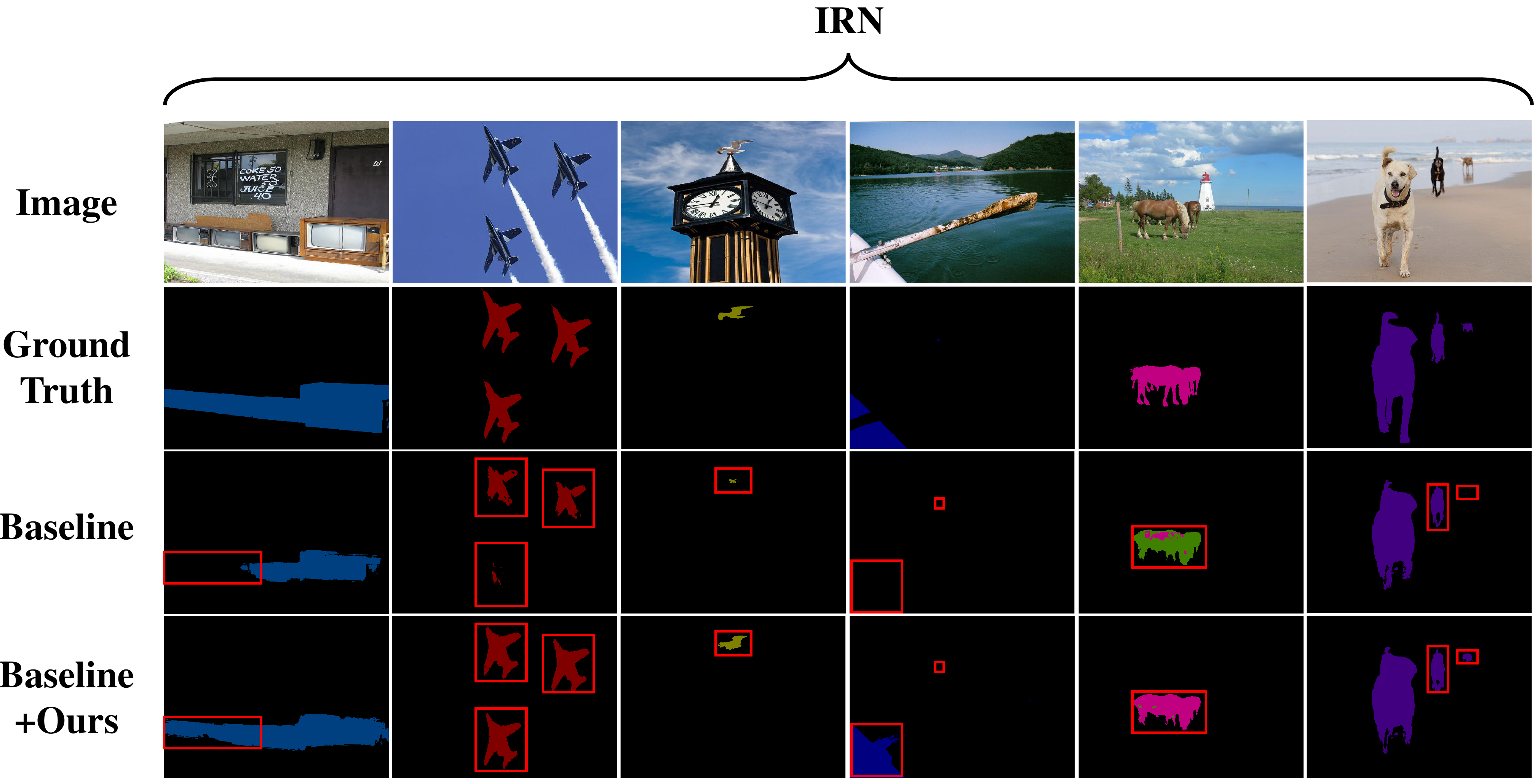}
        \caption{Visualization of IRN on PASCAL-B. Our loss function successfully fine-tunes baseline model to improve the ability of capturing objects including small-sized ones which is expressed by red bounding boxes.}
    \label{fig:our_irn}
\end{figure*}

\begin{figure*}[t!]
    \centering
        \includegraphics[width=\linewidth]{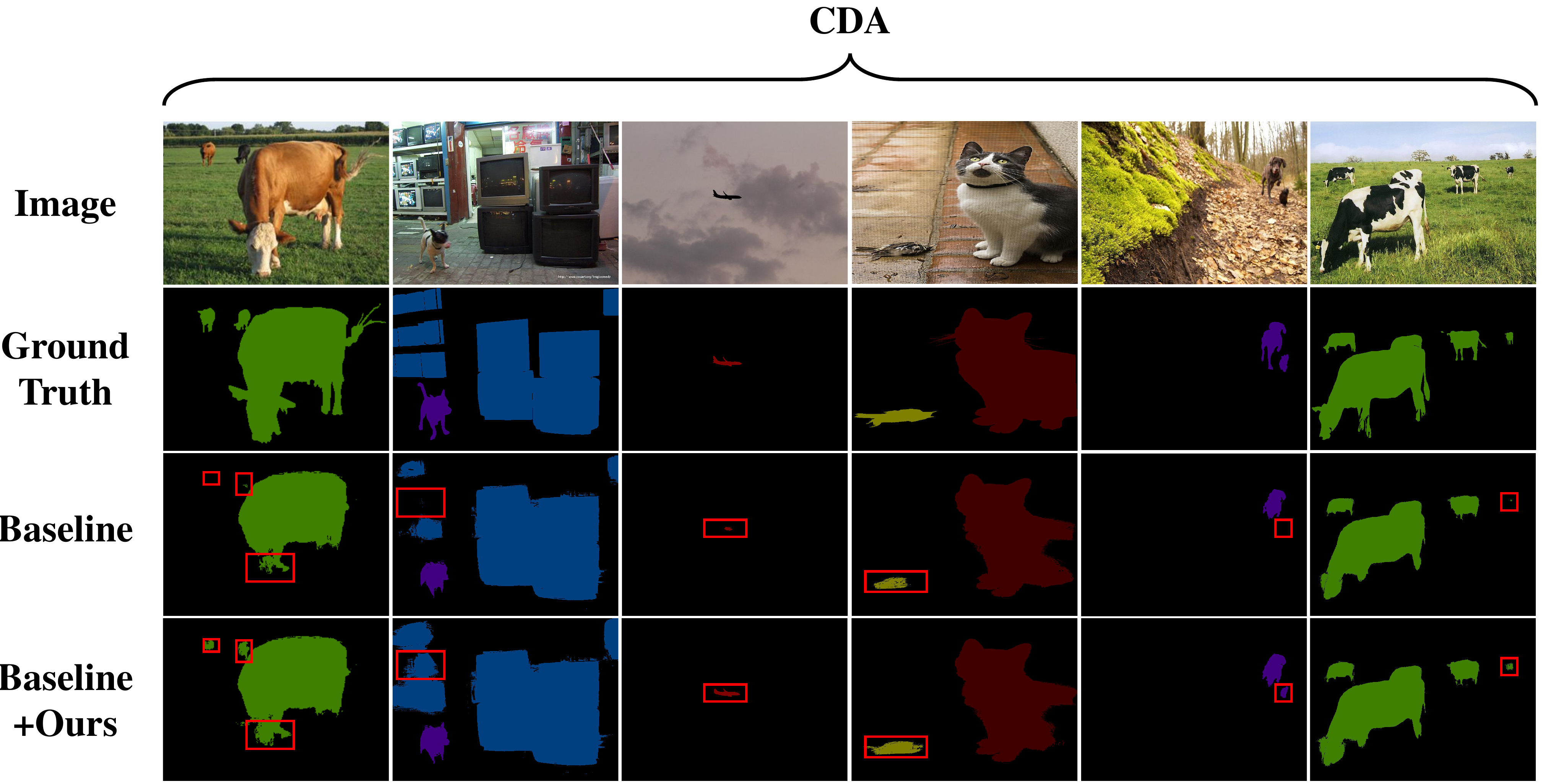}
        \caption{Visualization of CDA on PASCAL-B. Our loss function successfully fine-tunes baseline model to improve the ability of capturing objects including small-sized ones which is expressed by red bounding boxes.}
    \label{fig:our_cda}
\end{figure*}

\begin{figure*}[t!]
    \centering
        \includegraphics[width=\linewidth]{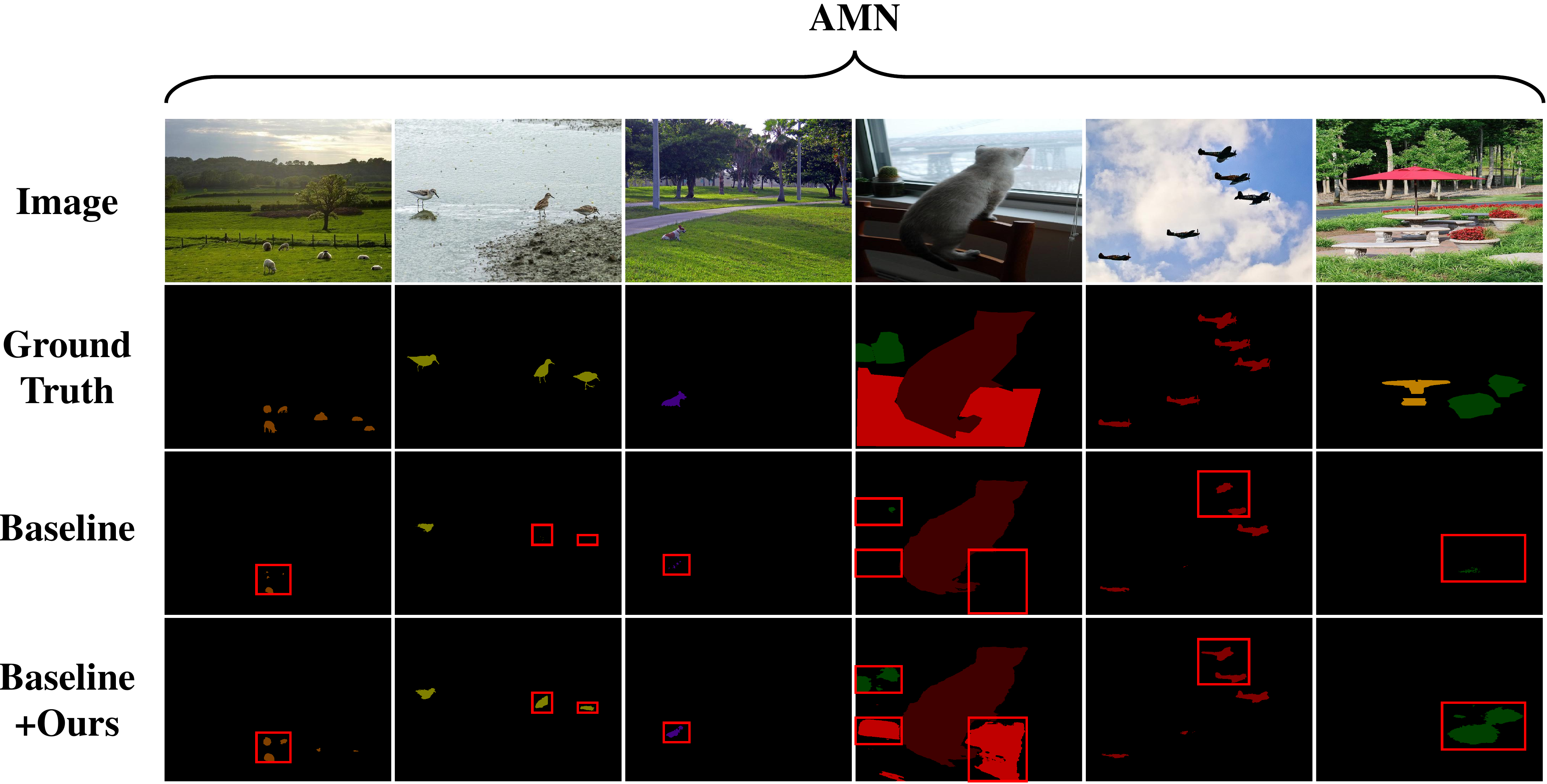}
        \caption{Visualization of AMN on PASCAL-B. Our loss function successfully fine-tunes baseline model to improve the ability of capturing objects including small-sized ones which is expressed by red bounding boxes.}
    \label{fig:our_amn}
\end{figure*}

\begin{figure*}[!t]
    \centering
        \includegraphics[width=\linewidth]{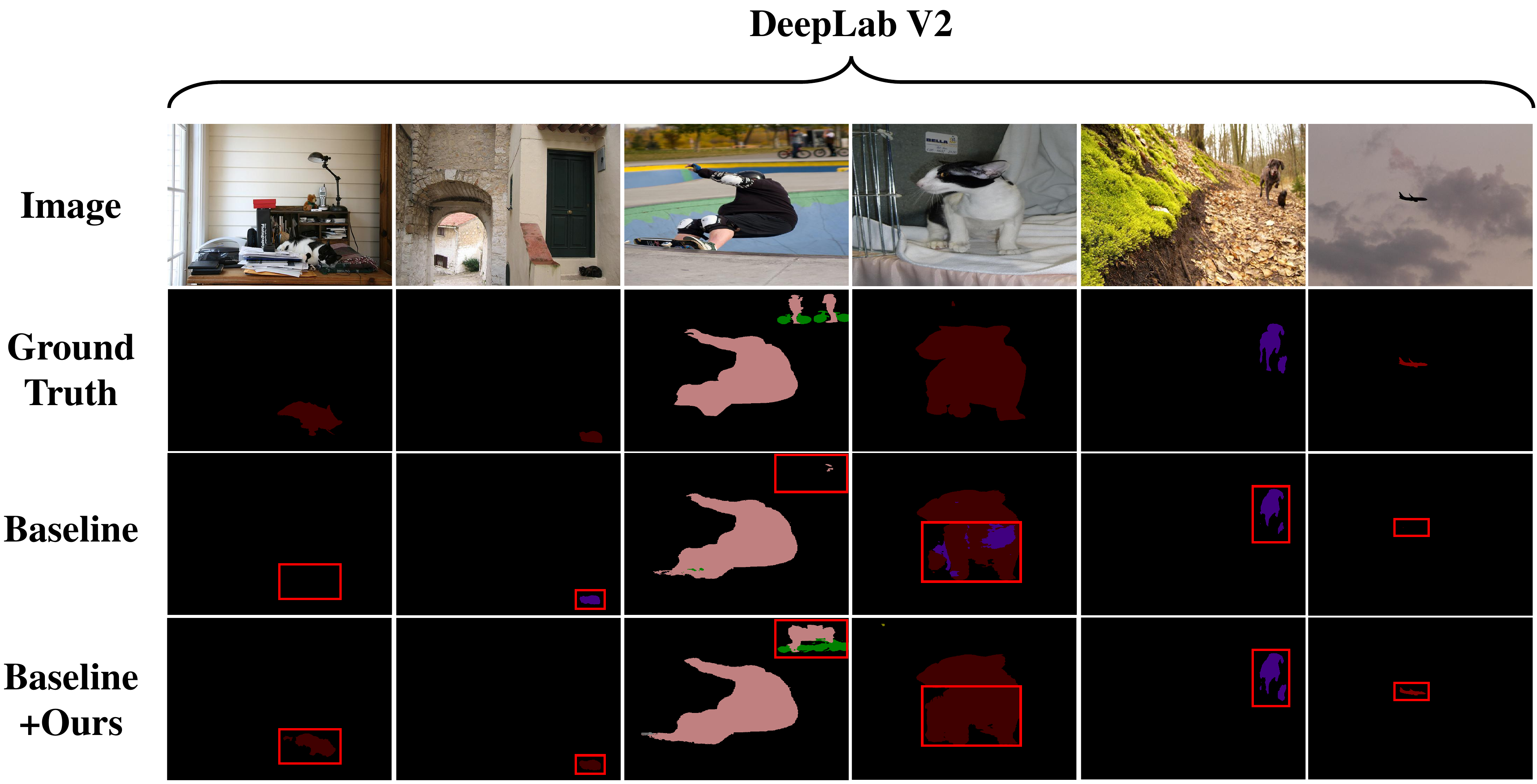}
        \caption{Visualization of DeepLab V2 on PASCAL-B. Our loss function successfully fine-tunes baseline model to improve the ability of capturing objects including small-sized ones which is expressed by red bounding boxes.}
    \label{fig:our_deeplabv2}
\end{figure*}

We show the visualization of prediction maps for each method~\cite{irn,rib,bbam,cda,bana,AMN,edam,Xie_2022_CVPR,nsrom,RCA,DeepLab} on three datasets: PASCAL VOC (from \fref{fig:voc_bbam} to~\fref{fig:voc_deeplab2}), MS COCO (from \fref{fig:coco_fully} to~\fref{fig:coco_rib}), and PASCAL-B (from \fref{fig:our_bbam} to~\fref{fig:our_deeplabv2}). Each figure shows that models with our loss function catch the objects more clearly including small-sized ones since our loss aims to constrain the network to be trained in balance considering the size of instances.

%%%%%%%%% REFERENCES
{\small
\bibliographystyle{ieee_fullname}
\bibliography{supplement}
}